\documentclass[10pt,journal,compsoc]{IEEEtran}
\usepackage{ctable}
\usepackage{caption}

\ifCLASSOPTIONcompsoc
  \usepackage[nocompress]{cite}
\else
  \usepackage{cite}
\fi
\ifCLASSINFOpdf
\else
\fi
\usepackage{graphicx}
\usepackage{float}
\usepackage{amssymb,amsfonts,amsmath,amsthm}

\usepackage{breqn}
\usepackage{algorithm,algpseudocode}
\usepackage{enumitem}

\usepackage[hidelinks]{hyperref}

\begin{document}
%
\title{A Complex Network based Graph Embedding Method for Link Prediction}
%
%
%
%

\author{Said~Kerrache, Hafida~Benhidour
\IEEEcompsocitemizethanks{\IEEEcompsocthanksitem King Saud University, College of Computer and Information Sciences, Riyadh, 11543, KSA.\protect\\
E-mail: hbenhidour@ksu.edu.sa}
}

\IEEEtitleabstractindextext{%
\begin{abstract}
Graph embedding methods aim at finding useful graph representations by mapping nodes to a low-dimensional vector space. It is a task with important downstream applications, such as link prediction, graph reconstruction, data visualization,  node classification, and language modeling. In recent years, the field of graph embedding has witnessed a shift from linear algebraic approaches towards local, gradient-based optimization methods combined with random walks and deep neural networks to tackle the problem of embedding large graphs. However, despite this improvement in the optimization tools, graph embedding methods are still generically designed in a way that is oblivious to the particularities of real-life networks. Indeed, there has been significant progress in understanding and modeling complex real-life networks in recent years. However, the obtained results have had a minor influence on the development of graph embedding algorithms. This paper aims to remedy this by designing a graph embedding method that takes advantage of recent valuable insights from the field of network science. More precisely, we present a novel graph embedding approach based on the popularity-similarity and local attraction paradigms. We evaluate the performance of the proposed approach on the link prediction task on a large number of real-life networks. We show, using extensive experimental analysis, that the proposed method outperforms state-of-the-art graph embedding algorithms. We also demonstrate its robustness to data scarcity and the choice of embedding dimensionality.
\end{abstract}

\begin{IEEEkeywords}
Graph embedding, link prediction, complex networks, popularity-similarity, local attraction, neural network.
\end{IEEEkeywords}}

\maketitle

\IEEEdisplaynontitleabstractindextext

%
\IEEEpeerreviewmaketitle

\IEEEraisesectionheading{\section{Introduction}\label{sec:introduction}}

Graph embedding aims at mapping the graph's nodes and edges into elements of a low-dimensional vector space while preserving its structural properties as much as possible \cite{embeddingsurvey18}. It is an upstream task with important applications, including link prediction \cite{embeddingsurvey18, kazemi18, hmsm16}, data visualization and exploration \cite{vanDerMaaten2008, largevis16}, recommendation systems \cite{matfact09},  node classification \cite{nodeclass11, nodeclass16}, and language modeling \cite{mikolov13, transformers}. Several approaches for graph embedding exist in the literature. These include algebraic methods based on matrix decomposition, such as Locally Linear Embedding \cite{lle}, Laplacian Eigenmaps \cite{lapeigen}, and Matrix Factorization  \cite{matfact09} (also known as graph factorization in \cite{embeddingsurvey18, graph-fact13});  random walk methods, such as DeepWalk \cite{deepwalk}, LINE \cite{line} and Node2Vec \cite{node2vec}; and deep learning-based methods \cite{dngr, sdne}.
The quality of a graph embedding is measured by its capacity to faithfully preserve the graph topology. Hence, in addition to its intrinsic importance and applications in various fields \cite{huang2005link, li2009recommendation, liu2007predicting, martinez2017survey, kerrache20}, link prediction is arguably the primary task by which graph embedding algorithms are evaluated. Graph embedding-based link prediction is generally achieved by appending a reconstruction function to the embedding algorithm. This function can be a straightforward similarity measure, such as dot product or cosine similarity \cite{deepwalk}, or a more complex nonlinear function learned from data using a neural network, for instance. 

This paper presents a novel graph embedding approach that exploits two paradigms that have been successfully used for link prediction: popularity-similarity \cite{Serrano2008Self-similarityofComplexNetworksandHiddenMetricSpaces, boguna2009navigability, papadopoulos2012popularity, zou2014exploiting, kab16, kerrache20,  Boguna2021} and local attraction \cite{kerrache20}. The popularity-similarity paradigm stipulates that the formation of links is governed by popularity, which is the tendency of a node to form connections and the similarity between nodes. Local attraction is a specific form in which the local neighborhood affects the likelihood of link formation. The higher the local attraction index between two nodes, the more likely they connect. Unlike previous work where these quantities are estimated from the observed network topology, we construct a joint embedding that encodes them in nodes' coordinates. The output of the embedding algorithm is then fed to a feed-forward neural network trained to discriminate between connected and disconnected couples. We demonstrate through extensive experimental analysis that the proposed method outperforms state-of-the-art graph embedding algorithms and show its robustness to data scarcity and the choice of embedding dimensionality.

The content of this paper is organized as follows.
Related work is reviewed in Section \ref{sec:related-work}. Section \ref{sec:problem} presents a formal definition of the link prediction problem, emphasizing embedding-based link prediction.
Section \ref{sec:proposed} contains a detailed description of the proposed approach.
Performance evaluation demonstrating the effectiveness of the proposed approach is presented in Section \ref{sec:experiments}.
Finally, Section \ref{sec:conclusion} concludes the paper and hints at a number of future research directions.

\section{Related Work}
\label{sec:related-work}
We divide this literature review into two parts. First, we present link prediction methods in general, including traditional topological similarity methods and more advanced techniques such as probabilistic and popularity-similarity methods. We then focus on methods based on graph embedding, where we discuss how an embedding can be used for link prediction and present the most important graph embedding techniques.

\subsection{Link prediction methods}
In this section, we will present the most important link prediction methods found in the literature. We will focus on methods based solely on topological information. Semantic approaches \cite{Hasan2006LinkPredictionUsingSupervisedLearning, Wang2007LocalProbabilisticModelsforLinkPrediction}, that is, those that use extra information associated with nodes, are out of the scope of this work. There is no universally agreed-upon taxonomy for link prediction methods, but for the purpose of this work, we will divide them into three main classes: topological similarity methods, probabilistic methods, and popularity-similarity methods. We postpone the discussion of embedding-based approaches until the next section.

Topological similarity measures are simple methods that use local topological information to predict missing links \cite{Liben-Nowell2007Link-predictionProblemforSocialNetworks}. Different topological measures have been proposed in the literature, including  common neighbors \cite{newman2001clustering} index, Jaccard's index \cite{jaccard1901etude}, preferential attachment \cite{newman2001clustering}, hub promoted Index \cite{ravasz2002hierarchical}, and Adamic and Adar index \cite{adamic2003friends} . 
The effectiveness  of these measures and their low computational requirements motivated the introduction of new measures later on, including resource allocation \cite{yang2015EvaluatingLinkPredictionMethods}, and Cannistraci-Ressource-Allocation  rule \cite{Cannistraci2013Fromlink-predictioninBrainConnectomesandProteinInteractomestothelocal-community-paradigmincomplexnetworks,daminelli2015common}. Combining several measure using an information-theoretical framework was also investigated in \cite{zhu2015information}, where the weight of each measure is chosen proportional to the value of the information it contributes to determining the existence of a link.

Unlike topological similarity measures, where topological information is directly used to score edges, probabilistic methods first build a probabilistic model to compute the connection probability between nodes. The Hierarchical Random Graph model introduced in \cite{Clauset2008HierarchicalStructureandthePredictionofMissingLinksinNetworks} consists of a binary tree where internal nodes represent nested clusters, whereas leaf nodes represent the vertices of the network. A probability is associated with each internal node representing the likelihood of the existence of a link between any two of its children. Hence, the probability assigned to the lowest common ancestor of any two nodes corresponds to their connection probability. 
In the Stochastic Block Model \cite{Guimer2009MissingandSpuriousInteractions}, the graph is partitioned into blocks where the probability of connection between any two nodes depends on the blocks to which they belong. The resulting model captures the community structure within the graph and can estimate link reliability, which allows to predict missing links and detect spurious ones. Although Hierarchical Random Graph model and Stochastic Block Model produce good results in general, they suffer from a high computational cost. Fast Blocking probabilistic  Model  \cite{Liu2013CorrelationsBetweenCommunityStructureAndLinkFormationInComplexNetworks} is a faster probabilistic model that uses a greedy strategy to estimate link probabilities based on link densities within and among communities. 

Popularity-similarity methods assume that the likelihood of a link between two nodes depends on the similarity between the nodes and their popularity. Similar nodes tend to connect with a high probability, and so do popular nodes. In \cite{Boguna2007NavigabilityofComplexNetworks}, the authors introduce a generative popularity-similarity model that uses a hidden metric space to encode nodes similarity. Nodes located at a smaller distance are considered more similar than those at a larger distance.  In \cite{Boguna2010SustainingtheInternetwithHyperbolicMapping}, the underlying metric space is assumed to possess a hyperbolic geometry, and in \cite{Papadopoulos2015NetworkMappingbyReplayingHyperbolicGrowth}, the authors present a method for embedding the graph in hyperbolic space and use the resulting embedding for link prediction. 
Authors in \cite{kerrache20} assume that the likelihood of links depends on the popularity of the nodes, their similarity, and the local attraction the nodes are subject to when they share the same neighborhood. The authors estimate the popularity and local attraction from the observed network topology and use them to pre-weight the graph. The similarity between nodes is calculated via shortest path distances based on the aforementioned weights.

\subsection{Graph embedding methods}
\label{sec:lr-embedding}
Graph embedding methods can be divided according to the underlying technique used to collect and encode the information on the graph topology and the models and algorithms used to find the embedding into three main categories: matrix factorization methods, random walk methods, and deep learning methods \cite{embeddingsurvey18}. In the rest of this section, we will survey each category and present the most prominent approaches therein.

\subsubsection{Factorization methods}
Factorization methods first capture the graph's topological information in a matrix, such as the adjacency matrix, the Laplacian matrix, or the transition matrix. The obtained matrix is then factorized to obtain an embedding. This step often involves a linear algebra decomposition algorithm, such as eigenvalue decomposition. In situations where algebraic tools cannot be employed, gradient-descent type algorithms can be used instead to find the embedding by optimization.

Locally Linear Embedding (LLE) \cite{lle} assumes that the network can be embedded in a Riemannian manifold, which implies that the geometry of the embedding is locally flat. Hence,  the coordinates of every node can be approximately written as a linear combination of its neighbors' coordinates:
\begin{equation}
	x^i = \sum_{j \in \Gamma_{i}} w_{ij} x^j,
\end{equation}
where $ \Gamma_{i} $ is the set of neighbors of $ i $. The entries of the network adjacency matrix can be used as values for the weights $ w_{ij} $. Computing the embedding can then be achieved by minimizing the following loss function:
\begin{equation}
	\sum_{i=1}^{n} \| x^i - \sum_{j \in \Gamma_{i}} w_{ij} x^j \|^2.
\end{equation}
Note that this problem admits the degenerate solution $ x^i=0, \forall i $. To eliminate this solution, the following constraint on the variance of the embedding is imposed: $ \frac{1}{n} X^T X= I$, where $ X $ is the matrix composed of all nodes coordinates $ x^i $. Additionally, the coordinates are centered at 0 to handle translational symmetry. Although this problem can be solved using gradient-descent-type algorithms, a more common approach is translating it into an eigenvalue problem. Let $ W $ be matrix composed of the weights $ w_{ij} $. It then can be shown that the solution to the above problem is the eigenvectors of the matrix $(I-W)^T(I-W)$ corresponding to eigenvalues ranked $ 2 \ldots d+1 $ when taken in ascending order; The eigenvector corresponding to the smallest eigenvalue is discarded.

Laplacian Eigenmaps \cite{lapeigen} aims at minimizing the distance between couples of nodes that have a high associated weight $ w_{ij} $:
\begin{equation}
\label{eq:lem}
\min_{x^i} \sum_{i, j} \dfrac{1}{2} w_{ij} \| x^i - x^j \|^2.
\end{equation}
The weights can be chosen using various approaches, the simplest of which is to assign the weight 1 to connected couples and 0 to disconnected couples. This problem can be shown to be equivalent to minimizing the trace of the matrix  $ X^T L X $, where $ L $ is the graph Laplacian. To avoid trivial solutions, the constraint $ X^T D X = I$, where $D$ is the diagonal matrix where $D_{i i}$ is the degree of node $i$. Similar to LLE, the minimization problem associated with Laplacian Eigenmaps can be reduced to an eigenvalue problem. It can be shown that the solution to \eqref{eq:lem} consists of the eigenvectors corresponding to the smallest eigenvalues of the normalized Laplacian matrix defined as $D^{-\frac{1}{2}} L D^{\frac{1}{2}}$, ignoring the eigenvalue zero.

Matrix factorization  \cite{matfact09} (also referred to as graph factorization in \cite{embeddingsurvey18, graph-fact13}) finds an embedding by solving
\begin{equation}
\label{eq:matfact}
\min_{x^i} \dfrac{1}{2} \sum_{(i,j) \in E} \| w_{i j } - x^i\cdot x^j \|^2 + \dfrac{\lambda}{2}  \sum_{i=1}^{n} \| x^i \|^2.
\end{equation}
Other factorization methods include Structure Preserving Embedding (SPE) \cite{spe2009}, Cauchy Graph Embedding \cite{cge2011} GraRep \cite{grarep2015} and HOPE \cite{hope2016}.

\subsubsection{Random walk methods}
Factorization methods can be impractical for very large graphs or graphs that are only partially observable. In such situations, random walk methods offer a more computationally affordable solution. One of the well known algorithms in this category is DeepWalk \cite{deepwalk}. It performs a set of random walks of length $2 w +1$ and finds an embedding map $\Phi$ that maximizes the probability that nodes $v_{i-w}, \ldots, v_{i-1}$ appear before node $v_i$ and nodes $v_{i+1}, \ldots, v_{i+w}$ appear after $v_i$. The algorithm uses dot product similarity to reconstruct the edges from the embedding and solves the embedding problem by minimizing the following negative log-likelihood over all walks:
\begin{equation}
\min_{\Phi} -\log p\left(v_{i-w}, \ldots, v_{i-1},v_{i+1}, \ldots, v_{i+w}\right | \Phi(v_i))
\end{equation}

Node2Vec \cite{node2vec} follows a similar procedure to DeepWalk but uses a biased random walk that offers a balance between breadth first and depth first traversals around the node, providing hence a more accurate description of the node's neighbourhood. The algorithm requires the computation of a per-node partition function which is estimated using negative sampling \cite{mikolov13}. The embedding vectors are found using stochastic gradient descent.

Hierarchical Representation Learning for Networks (HARP) \cite{harp} uses a multi-resolution approach to mitigate the issue of convergence to local minima encountered with DeepWalk and Node2Vec. HARP creates a hierarchy of graphs, each obtained from the graph in the level below by grouping nodes together using graph coarsening techniques. The embedding process is then started at the top of the hierarchy. Once embedded, the node vectors obtained at level $i$ are used as initialization for the vectors at level $i+1$. The emebddings are thus propagated downwards until the deepest level corresponding to the original graph. This allows to find better minima for the objective functions of DeepWalk and Node2Vec for instance.

There also exist several methods that improve on the above approaches or combine them with factorization methods \cite{walklet,ddrw,tridnr,Yang2016RevisitingSL,genvector}.

\subsubsection{Deep learning and other methods}
Structural Deep Network Embedding method (SDNE) \cite{sdne} uses a semi-supervised deep model to capture the non-linear network structure. The model has two components: an unsupervised component that uses the second-order proximity to capture the global network structure and a supervised component used to preserve first-order proximity and hence the local network structure.
Deep Neural Networks for Learning Graph Representations (DNGR)
\cite{dngr} starts by using a random surfing model to generate a co-occurrence matrix, which is then transformed into a positive
mutual information matrix and fed into a stacked denoising autoencoder. The autoencoder allows for capturing non-linearities and extracting complex features.

Graph Convolutional Networks  (GCNs) extend the idea of Convolutional Neural Networks used with images to graphs. Similar to CNNs, GCNs define convolution operators acting on the nodes' local neighborhoods. This process is repeated iteratively over several layers aggregating in the process information from increasingly larger neighborhoods. Unlike SDNE and DNGR, the iterative application of local convolution operators makes GCN scalable to large graphs and allows them to capture complex topological structures \cite{gcn}.

LINE \cite{line} works on two proximity levels, first-order proximity, which reflects direct node-to-node connection, and second-order proximity, which is concerned with indirect connections through common neighbors. For the first-order proximity, LINE defines two probability functions, one derived from the adjacency matrix and the other from the embedding. The algorithm then finds the best embedding by minimizing the Kullback-Leibler divergence of these two probability distributions. A similar strategy is used with second-order proximity. The final embedding of a node is the concatenation of its first-order and second-order embeddings.

\subsection{Discussion}
Topological similarity measures have been widely successful in solving the link prediction problem. They are simple to understand and implement, and unlike probabilistic methods, for instance, they are highly computationally efficient and hence scalable to massive networks. This efficiency comes at a price, however. Because they rely on local information, topological similarity measures perform poorly when data is scarce, such as in cold start situations. Furthermore, these methods give little insight into the structure and function of the network since their output is solely intended for link scoring. Graph embedding methods not only solve the link prediction problem robustly but also generate a valuable network representation that can be used with various other downstream tasks, such as community detection, data visualization, and node classification. 

In recent years, graph embedding has shifted from linear algebra-based methods to local, gradient-based optimization methods combined with random walks or deep neural networks to tackle the problem of embedding large graphs. This tendency is primarily influenced by the enormous success of gradient-based methods, particularly stochastic gradient methods, in complex model learning, notably deep neural models \cite{GoodfellowIan2016Dl}. Despite this improvement in the optimization tools, graph embedding methods are still designed in a very generic way that is oblivious to the specificities of networks they target. During the last two decades, there has been a substantial research effort within the field of network science to understand and model the structure and function of real-life networks, and considerable progress has been made in that direction \cite{Albert2002StatisticalMechanicsofComplexNetworks, Newman2003StructureandFunctionofComplexNetworks, Boguna2021}. However, the results from the field of network science have had a little influence on the research on graph embedding. This work aims to design a graph embedding method that takes advantage of recent valuable insights on the topology of complex networks. More precisely, we aim to develop an embedding algorithm based on the popularity-similarity \cite{ Boguna2007NavigabilityofComplexNetworks,Serrano2008Self-similarityofComplexNetworksandHiddenMetricSpaces, Boguna2021} paradigm combined with local attraction \cite{kerrache20}. The premise is that taking into consideration the particularities of real-life networks can lead to better embedding algorithms.

\section{Problem Statement}
\label{sec:problem}
Consider a graph $G(V, E)$, where $V$ is the set of nodes, and $E$ is the set of edges. The set $E$ is also referred to as the set of \emph{positive links}, and its complement  $\bar{E}$ with respect to the set of all possible edges on $G$  is referred to as the set of \emph{negative links}. The link prediction problem is determining which elements of $\bar{E}$ are more likely to be missing from the network or, in the case of an evolving network, may appear shortly \cite{Lu2011LinkPredictioninComplexNetworks}. Link prediction algorithms assign scores to non-existing edges, with the convention that the higher score, the more likely the corresponding link is missing. 

Link prediction by graph embedding consists in finding a mapping $\Phi: V \to \mathbb{R}^d$ that assigns coordinates $x^i \in \mathbb{R}^d$ to every node $i \in V$ and a reconstruction function $\Psi: \mathbb{R}^d \times \mathbb{R}^d \to \mathbb{R}$ used to score non-existing links based solely on their coordinates. More likely edges should be assigned higher scores than unlikely ones. The reconstruction function can be hand-designed or learned in a supervised manner from the observed network topology.

\section{Proposed approach}
\label{sec:proposed}
We propose a network embedding approach that combines two types of embeddings. The first one follows the popularity-similarity paradigm and attributes the observed network topology to two properties of the nodes, their popularity, represented by their connection degree, and their similarity as reflected by their hidden coordinates. The second embedding explains the connections between nodes by the local attraction resulting from sharing the same neighborhood. We propose two approaches to combine these embeddings. The first approach uses a single objective function that merges the two terms, popularity-similarity and local interaction. On the other hand, the second approach independently fits the two models to simplify the resulting optimization problem. The final embedding is the concatenation of the coordinates obtained from the two separate embeddings.

\subsection{Embedding based on popularity-similarity}
\label{sec:ps}
Unlike the hidden metric formulation used in \cite{hmsm16, Boguna2007NavigabilityofComplexNetworks,Serrano2008Self-similarityofComplexNetworksandHiddenMetricSpaces}, we model similarity between nodes as the dot product between nodes coordinates scaled by node popularity. This linear form results in a simpler optimization problem compared to the metric model, and therefore allows the approach to scale to larger networks.

We define the normalized popularity of node $ i $ as:
\begin{equation}
	\label{eq:npop}
	\pi_i =  \dfrac{\log(\kappa_i+2)}{\log(\kappa_{\max}+2)},
\end{equation}
where $ \kappa_i $ is the degree of node $ i $, and $ \kappa_{\max} $ is the maximum degree in the network.
We assume that the probability of the existence of a link between two nodes $ i, j $ given their scaled coordinates $ \Tilde{x}^i, \Tilde{x}^j \in \mathbb{R}^{d}$ and their normalized popularity $ \pi_i, \pi_j $, correlates positively with the score $ \psi_{i j} $ defined as follows:
\begin{equation}
	\label{eq:ps}
	\psi_{i j} = \pi_i \pi_j \Tilde{x}^i \cdot \Tilde{x}^j = \pi_i \pi_j \sum_{k=1}^{d} \Tilde{x}^i_k \Tilde{x}^j_k.
\end{equation}
The full embedding of a node requires the inclusion of its popularity, that is $ \Phi_{ps}(i)=x^i=\pi_i \Tilde{x}^i \in \mathbb{R}^d$. 

To determine the hidden coordinates, we assign a pre-defined score to positive links, which we denote by $ \psi_1 $, and a score $ \psi_0 $ to negative links. Typically, we set $ \psi_1 \approx 1$, whereas $ \psi_0 \approx 0$. We then minimize the following objective function:

\begin{multline}
	\label{eq:ps-l2}
	J^{L_2}_{ps}\left(\Tilde{x}^1,\ldots,\Tilde{x}^n\right)=\dfrac{1}{2} \sum_{(i,j) \in E} \left(\Tilde{x}^i \cdot \Tilde{x}^j - \dfrac{\psi_1}{\pi_i \pi_j} \right)^2\\ + \dfrac{1}{2} \sum_{(i,j) \notin E} \left(\Tilde{x}^i \cdot \Tilde{x}^j - \dfrac{\psi_0}{\pi_i \pi_j} \right)^2 + \dfrac{\lambda}{2}  \sum_{i=1}^{n} \| \Tilde{x}^i \|_2^2,
\end{multline}
where $ \lambda $ is the regularization coefficient, and $\| \cdot \|_2$ stands for the $L_2$ norm. 
Alternatively, we can use the $L_1$ norm for the error cost and regularization:
\begin{multline}
	\label{eq:ps-l1}
	J^{L_1}_{ps}\left(\Tilde{x}^1,\ldots,\Tilde{x}^n\right)= \sum_{(i,j) \in E} \left|\Tilde{x}^i \cdot \Tilde{x}^j - \dfrac{\psi_1}{\pi_i \pi_j} \right|\\ + \sum_{(i,j) \notin E} \left|\Tilde{x}^i \cdot \Tilde{x}^j - \dfrac{\psi_0}{\pi_i \pi_j} \right| + \lambda  \sum_{i=1}^{n} \| \Tilde{x}^i \|_1,
\end{multline}
where $ \| \cdot \|_1$ represents the $L_1$ norm of the vector. 

The number of terms in the least-squares part of \eqref{eq:ps-l2} and \eqref{eq:ps-l1} is equal to $ n\left(n-1\right)/2 $, $ n $ being the number of nodes. This can become prohibitively big for large networks. In that case, we may resort to sub-sampling to reduce the complexity of the resulting optimization problem. It is possible to sub-sample both positive and negative links, but since the number of negative links typically dominates this term, we will use all positive links and only sub-sample the negative links.

\subsection{Embedding based on local attraction}
The observation that nodes in the same neighborhood tend to connect is at the basis of several link prediction algorithms, particularly local methods. In this paper, we follow \cite{kerrache20}, and assume that the common neighborhood between two nodes causes an attraction between nodes. More precisely, two nodes $i$ and $j$ experience an attraction $ \eta_{i j} $ given by \cite{kerrache20}:
\begin{equation}
\eta_{ij} = 1 - \prod_{k \in \Gamma_{i j}} \dfrac{\log(\kappa_{k}+1)}{\log(\kappa_{\max}+1)}, 
\end{equation}
where $\Gamma_{i j}$ is the set of common neighbors of $ i $ and $j$, and $ \kappa_k $ is the degree of node $ k $. By convention $ \eta_{ij} $ is set to 0 if $\Gamma_{i j}$ is empty.

In this paper, we assume further that this attraction factor can be expressed as the dot product between the nodes coordinates:
\begin{equation}
	\label{eq:attraction-hidden}
	\eta_{ij} = x^i \cdot x^j = \sum_{k=1}^{d} x^i_k x^j_k.
\end{equation}
Therefore, given the observed network topology, we can estimate nodes coordinates using a least squares principle by minimizing the following objective function:
\begin{multline}
\label{eq:la-l2}
	J^{L_2}_{la}\left(x^1,\ldots,x^n\right)=\dfrac{1}{2} \sum_{i=1}^{n} \sum_{j=i+1}^{n} \left(x^i \cdot x^j - \eta_{i j}\right)^2 \\+ \dfrac{\lambda}{2}  \sum_{i=1}^{n} \| x^i \|_2^2.
\end{multline}
where $ \lambda $ is the regularization coefficient. Of course, the $L1$ norm can also be used instead:
\begin{multline}
\label{eq:la-l1}
	J^{L_1}_{la}\left(x^1,\ldots,x^n\right)= \sum_{i=1}^{n} \sum_{j=i+1}^{n} \left|x^i \cdot x^j - \eta_{i j}\right| \\+ \lambda \sum_{i=1}^{n} \| x^i \|_1.
\end{multline}

Similarly to what was discussed in the previous section, we may resort to sub-sampling to reduce the computational computational complexity of this optimization problem.

\subsection{Combining the embeddings}
The first approach to combine popularity-similarity and local attraction is to embed the network using the two approaches separately and then simply concatenate the two resulting vectors: $\Phi_{psl}(i)=\left(\Phi_{ps}(i)^T, \Phi_{la}(i)^T\right)^T$. Concatenation simplifies the optimization since it divides the problem into two smaller and thus simpler optimization tasks.

Alternatively, we can merge all three factors into a single combined optimization problem. The basic idea is to push couples with a high local attraction to have higher scores than couples with a low or null local attraction, which can, for instance, be achieved by minimizing the following objective function:
\begin{multline}
	\label{eq:psl-l2}
	J^{L_2}_{psl}\left(\Tilde{x}^1,\ldots,\Tilde{x}^n\right)=\dfrac{1}{2} \sum_{(i,j) \in E} \left(\Tilde{x}^i \cdot \Tilde{x}^j - \dfrac{\psi_1 (1 + \eta_{ij})}{\pi_i \pi_j} \right)^2\\ + \dfrac{1}{2} \sum_{(i,j) \notin E} \left(\Tilde{x}^i \cdot \Tilde{x}^j - \dfrac{\psi_0 (1 + \eta_{ij})}{\pi_i \pi_j} \right)^2 + \dfrac{\lambda}{2}  \sum_{i=1}^{n} \| \Tilde{x}^i \|_2^2,
\end{multline}
where $\psi_1$ and $\psi_0$ are defined as in Section \ref{sec:ps}. If the $L_1$ norm is used instead, we obtain the following objective:
\begin{multline}
	\label{eq:psl-l1}
	J^{L_1}_{psl}\left(\Tilde{x}^1,\ldots,\Tilde{x}^n\right)= \sum_{(i,j) \in E} \left|\Tilde{x}^i \cdot \Tilde{x}^j - \dfrac{\psi_1 (1 + \eta_{ij})}{\pi_i \pi_j} \right|\\ + \sum_{(i,j) \notin E} \left|\Tilde{x}^i \cdot \Tilde{x}^j - \dfrac{\psi_0 (1 + \eta_{ij})}{\pi_i \pi_j} \right| + \lambda \sum_{i=1}^{n} \| \Tilde{x}^i \|_1.
\end{multline}
The multiplication by $1+\eta_{ij}$ instead of $\eta_{ij}$ avoids forcing couples with no common neighbors to be disconnected.
Here also, the full embedding of a node is defined as $ \Phi_{psl}(i)=x^i=\pi_i \Tilde{x}^i \in \mathbb{R}^d$. 

\section{Performance evaluation}
\label{sec:experiments}
We perform several experiments on real-life networks to evaluate the proposed method's performance. We start by investigating the performance of different variants of the proposed approach and compare various design choices. We then compare the performance of our approach to state-of-the-art embedding methods through extensive experimental evaluation. To gain a deeper understanding of the proposed approach, we experimentally investigate the effect of embedding dimensionality and robustness under various degrees of data scarcity.

Before delving into the experimental results, we present the datasets and performance measures used in the evaluation, followed by a discussion of the implementation of the proposed approach and competing methods.

\subsection{Data}
We use 61 real-life networks available on public data repositories \cite{snapnets,  konect, pajaknets}. The networks used vary in size and type, including biological networks, citation networks, ecological networks, communication networks, transportation networks, and social networks. We divide the networks according to their size into two categories. The first category, shown in Table \ref{tab:data}, contains 40 small networks containing less than 1000 nodes. The size of these networks allows for conducting extensive experiments to test different design decisions and comparison scenarios. Larger networks shown in Table \ref{tab:large} are used as the conclusive comparison task.

\begin{table*}[!t]
	\centering 
	\caption{Description of the networks with less than 1000 nodes used in the experimental performance analysis. Columns $n$ and $m$ represent the number of nodes and links in the network, respectively.}
	\label{tab:data}
	\begin{tabular}{l p{11cm} r  r } 
		\toprule
		Network & Description & $n$ & $m$ \\
		\midrule
ACM2009 Contacts\cite{crowd11} & Face-to-face contact network of the attendees of the ACM Conference on Hypertext and Hypermedia 2009. Source: \url{http://konect.cc/networks/sociopatterns-hypertext} & 113 & 2,196 \\ \midrule

C. Elegans Metabolic \cite{Jeong2000LargeScaleOrganizationOfMetabolicNetworks} & Metabolic network of the worm C. Elegans \cite{Brenner1974GeneticsofCaenorhabditisElegans}. Source: \url{http://konect.cc/networks/arenas-meta}  & 453 & 2,038 \\ \midrule

C. Elegans Neural \cite{watts1998collective, white-1986-cneural} & A symmetrized version of the neural network of the the worm C. Elegans. Source: \url{http://cdg.columbia.edu/cdg/datasets} & 297 & 2,148 \\ \midrule

Centrality Literature \cite{centrality-90} & Network centrality citation network from 1948 to 1979. Source: \url{http://vlado.fmf.uni-lj.si/pub/networks/data/GD/a01.zip} &118 & 613 \\ \midrule

Chesapeake Lower \cite{chesapeake2002} & Lower Chesapeake Bay food web in summer. Source: \url{http://vlado.fmf.uni-lj.si/pub/networks/data/bio/foodweb/ChesLower.paj} & 37 & 167 \\ \midrule

Chesapeake Middle \cite{chesapeake2002} & Middle Chesapeake Bay food web in summer. Source: \url{http://vlado.fmf.uni-lj.si/pub/networks/data/bio/foodweb/ChesMiddle.paj} &37 &198 \\ \midrule

Chesapeake Upper \cite{chesapeake2002} & Upper Chesapeake Bay food web in summer. Source: \url{http://vlado.fmf.uni-lj.si/pub/networks/data/bio/foodweb/ChesUpper.paj} & 37 & 199 \\ \midrule

Codeminer \cite{Heymannm2008Javacode} & The call-graph of Java program. Node are packages, classes, fields and methods. Edges represent calls and containments. Source: \url{https://github.com/gephi/gephi.github.io/tree/master/datasets} & 724 & 1,015 \\ \midrule

CPAN Authors & Network of relationships between the developers of the Perl language. An edge between two nodes indicates that the two corresponding developers use the same Perl module. Source: \url{https://gephi.org/datasets/cpan-authors.gexf.zip} & 839 & 2,112 \\ \midrule

Cypress Dry\cite{florida2000} & Food web of Cypress (Florida) in the dry season. Source: \url{http://vlado.fmf.uni-lj.si/pub/networks/data/bio/foodweb/cypdry.paj} & 71 & 618 \\\midrule

Cypress Wet\cite{florida2000} & Food web of Cypress (Florida) in the wet season. Source: \url{http://vlado.fmf.uni-lj.si/pub/networks/data/bio/foodweb/cypwet.paj} & 71 & 612 \\\midrule

DNA Citation \cite{Hummon1989ConnectivityInCitationNetworkDNA} & DNA research literature citation network. Source: \url{http://vlado.fmf.uni-lj.si/pub/networks/Data/cite/default.htm}.  &39 &61\\ \midrule

DNA Citation CC\cite{Hummon1989ConnectivityInCitationNetworkDNA} & The main connected component of the DNA Citation network. Source: \url{http://vlado.fmf.uni-lj.si/pub/networks/Data/cite/default.htm} & 35 & 59 \\ \midrule

E.Coli\cite{shen2002network} & The transcriptional regulation network of Escherichia coli. Source: \url{http://vlado.fmf.uni-lj.si/pub/networks/data/GD/GD.htm} & 418 & 519 \\ \midrule

Erdos 971 & The 1971 version of Erd\"{o}s’ co-authorship network. Source: \url{https://sparse.tamu.edu/MM/Pajek/Erdos971.tar.gz} & 433 & 1,314 \\ \midrule

Erdos 981 & The 1981 version of Erd\"{o}s’ co-authorship network. Source: \url{https://sparse.tamu.edu/MM/Pajek/Erdos981.tar.gp{9.5cm}z} & 445 & 1,381 \\ \midrule

Erdos 991 & The 1991 version of Erd\"{o}s’ co-authorship network. Source: \url{https://sparse.tamu.edu/MM/Pajek/Erdos991.tar.gz} & 454 & 1,417 \\ \midrule

Everglades\cite{florida2000} & Food web of Everglades graminoid marshes in the wet season. Source: \url{http://vlado.fmf.uni-lj.si/pub/networks/data/bio/foodweb/Everglades.paj} & 69 & 880 \\ \midrule

GD 01 & Citation network of the Graph Drawing (GD) context from GD94 to GD2000. Source: \url{http://vlado.fmf.uni-lj.si/pub/networks/data/GD/a01.zip} & 259 & 640 \\ \midrule

Haggle Contact\cite{haggle07} & Contact network between persons measured by carried wireless devices. Source: \url{http://konect.cc/networks/contact} & 274 & 2,124 \\ \midrule

Infectious\cite{crowd11} & Face-to-face interaction network between visitors of the exhibition INFECTIOUS: STAY AWAY in 2009 at the Science Gallery in Dublin. Source: \url{http://konect.cc/networks/sociopatterns-infectious} & 410 & 2,765 \\ \midrule

Japan Air \cite{guimera2005} & Japanese air transportation network extracted from the World Transport network. Source: \url{http://seeslab.info/media/filer_public/63/97/63979ddc-a625-42f9-9d3d-8fdb4d6ce0b0/airports.zip} & 56 & 183 \\ \midrule

Jazz \cite{gleiser2003-jazz} & Jazz musicians collaboration network. Source: \url{http://deim.urv.cat/~alexandre.arenas/data/xarxes/jazz.zip} & 198 & 2,742 \\ \midrule

Les Miserables \cite{knuth93} & Character co-appearance network in the novel ''Les Miserables'' by Victor Hugo. Source: \url{http://www-personal.umich.edu/~mejn/netdata/lesmis.zip}. & 77 & 254 \\ \midrule

Macaque Neural \cite{Modha2010NetworkArchitectureMacaqueBrain} & The macaque brain network. Nodes represent brain regions, and edges correspond to long-distance connections between them. Source: \url{http://www.pnas.org/content/107/30/13485.full}. & 360 & 5,208 \\
\bottomrule
\end{tabular}
\end{table*}

\begin{table*}[!t]
\ContinuedFloat
\centering 
	\caption{(continued)}
	\begin{tabular}{l p{11cm} r  r } 
		\toprule
		Network & Description & $n$ & $m$ \\
		\midrule

Manufacturing e-mail\cite{michalski2011matching} & Email network within a manufacturing company. Source: \url{https://www.ii.pwr.edu.pl/~michalski/datasets/manufacturing.tar.gz}. &167 &3,250\\ \midrule

Maspalomas \cite{maspalomas} & Food web of Maspalomas coastal lagoon. Source: \url{http://vlado.fmf.uni-lj.si/pub/networks/data/bio/foodweb/Maspalomas.paj}. & 24 & 77 \\ \midrule

Narragan \cite{narragan} & Food web of the Narragansett estuary. Source: \url{http://vlado.fmf.uni-lj.si/pub/networks/data/bio/foodweb/Narragan.paj}. & 35 & 204 \\ \midrule

Physicians\cite{coleman1957diffusion} & Network of innovation spread among physicians in the towns in Illinois, Peoria, Bloomington, Quincy and Galesburg in 1966. Edges represent friendship or advice seeking between physicians. Source: \url{http://konect.cc/networks/moreno\_innovation}. & 241 & 923 \\ \midrule

Polbooks & Network of frequently co-purchased books on US politics sold by Amazon.com. Source: \url{http://www-personal.umich.edu/~mejn/netdata/}.  & 105 & 441 \\ \midrule

Political Blogs \cite{Adamic2005PoliticalBlogosphere} & Network of hyperlinks among blogs on US politics. Source: \url{http://networkrepository.com/web-polblogs.php}. & 643 & 2,280 \\ \midrule

Residence Hall\cite{residencehall98} & Friendship network among students living in a residence hall at the Australian National University campus. Source: \url{http://moreno.ss.uci.edu/data.html#oz}. &217 &1,839 \\ \midrule

School\cite{school11} & Face-to-face proximity network among students and teachers in a primary school. Source: \url{http://www.sociopatterns.org/datasets/primary-school-cumulative-networks} (day 1). & 236 & 5,899 \\ \midrule

SFBD Food Web\cite{florida2000} & South Florida food web in the dry season. Source: \url{http://vlado.fmf.uni-lj.si/pub/networks/data/bio/foodweb/baydry.paj}. & 128 & 2,106 \\ \midrule

SFBW Food Web\cite{florida2000} & South Florida food web in the wet season. Source: \url{http://vlado.fmf.uni-lj.si/pub/networks/data/bio/foodweb/baywet.paj}. & 128 & 2,075 \\ \midrule

StMarks \cite{stmarks} & Food web of St Marks National Wildlife Refuge. Source: \url{http://vlado.fmf.uni-lj.si/pub/networks/data/bio/foodweb/stmarks.paj}. & 54 & 350 \\ \midrule

Terrorist Train Bombing \cite{Hayes2006ConnectingtheDots} & Network of contacts among a set of terrorists involved in the Madrid train bombing of March 11, 2004. Source: \url{http://konect.cc/networks/moreno_train}. & 64 & 243 \\ \midrule

Terrorist\cite{krebs2002mapping} & Network of social associations among terrorists involved in the 9/11. Source: \url{http://tuvalu.santafe.edu/~aaronc/hierarchy/terrorists.zip}. & 62 & 152 \\ \midrule

US Air 97 & North American Transportation Atlas Data (NORTAD). Source: \url{http://vlado.fmf.uni-lj.si/pub/networks/data/map/USAir97.net}. &332 &2,126 \\ \midrule

Zakary's Karate Club \cite{Zachary1977InformationFlowModelforConflictandFissioninSmallGroups} & Network of friendships between members of a karate club at an American university. Source: \url{http://konect.cc/networks/ucidata-zachary} & 34 & 78 \\

\bottomrule
\end{tabular}
\end{table*}

\begin{table*}[!t]
	\centering 
	\caption{Description of the networks with more than 1000 nodes  used in the experimental performance analysis. Columns $n$ and $m$ represent the number of nodes and links in the network, respectively.}
	\label{tab:data-large}
	
	\begin{tabular}{l p{11cm} r  r } 
		\toprule
		Network & Description & $n$ & $m$ \\
		\midrule
		Adolescent Health\cite{adolescent-health} & Friendship network among adolescent students. Source: \url{http://konect.cc/networks/moreno_health}. &2,539 & 12,969\\ \midrule
		
		Advogato\cite{advogato} & The Advogato trust network. Source: \url{http://konect.cc/networks/advogato}. &5,155 &39,285\\ \midrule
		
		BitcoinAlpha\cite{bitcoinalpha} & The Bitcoin Alpha platform trust network. Only links with positive trust are considered. Source: \url{http://konect.cc/networks/soc-sign-bitcoinalpha}. &3,683 & 22,650\\ \midrule
		
		Ciao\cite{ciao} & Trust network from \url{http://dvd.ciao.co.uk} during 2013. Source: \url{http://konect.cc/networks/librec-ciaodvd-trust}. &4,658 & 40,133\\ \midrule
		
		Criminal\cite{criminal} & Network of phone calls among the members of a drug trafficking group. Source: \url{https://sites.google.com/site/ucinetsoftware/datasets/mainaseuropoldatasets}. &2,749 &2,952 \\ \midrule
		
		Diseasome \cite{diseasome} & A network of disorders and disease genes linked by disorder–gene associations. Source: \url{http://gephi.org/datasets/diseasome.gexf.zip}.  &1,419 &2,738 \\ \midrule
		
		DNC Email \cite{konect} &  Leaked emails network from the 2016 Democratic National Committee. Source: \url{http://konect.cc/networks/dnc-temporalGraph}. &1866 & 5517\\ \midrule
		
		FAA \cite{konect} & Network of preferred routes between US airports based on data from the National Flight Data Center (NFDC) of the USA's FAA (Federal Aviation Administration). Source: \url{http://konect.cc/networks/maayan-faa}. &1,226 &2,408 \\ \midrule
		
		Facebook \cite{facebook-12, facebook-14} & A Facebook friendship network. Source: \url{http://snap.stanford.edu/data/egonets-Facebook.html}. &4,039 & 88,234\\\midrule
		
		GR \cite{leskovec-2007} & Collaboration network among authors in arXiv's general relativity and quantum cosmology section. Source: \url{https://snap.stanford.edu/data/ca-GrQc.html}. &5,241 &14,484\\\midrule
		
		Hero  & A social network of the Marvel universe super heroes constructed by Cesc Rosselló, Ricardo Alberich, and Joe Miro from the University of the Balearic Islands. Source \url{https://gephi.org/datasets/hero-social-network.gephi}. &10,469	&178,115 \\ \midrule
		
		Human Protein\cite{human-protein} & Network of protein interaction in Humans. Source: \url{http://konect.cc/networks/maayan-Stelzl}. &1,702 & 6,171 \\ \midrule
		
		Indochina 2004\cite{bovwfi, brsllp} & A web network. Source: \url{http://networkrepository.com/web_indochina_2004.php}. &11,358 &47,606\\ \midrule
		
		ODLIS\cite{Reitz2002} & Network of cross-references in the Online Dictionary of Library and Information Science (ODLIS).  Source \url{http://vlado.fmf.uni-lj.si/pub/networks/data/dic/odlis/Odlis.htm}. &2,900 &16,377 \\ \midrule
		
		PGP \cite{Boguna2004ModelsofSocialNetworksBasedonSocialDistanceAttachment} & Network of users who exchange secure information using Pretty Good Privacy (PGP) algorithm. Source: \url{http://deim.urv.cat/~alexandre.arenas/data/welcome.htm}. & 10,680 & 24,316 \\ \midrule
		
		Roget \cite{roget62,knuth93} & A network of cross-references among the 1022 categories in Roget's Thesaurus. Source: \url{http://vlado.fmf.uni-lj.si/pub/networks/data/dic/roget/Roget.htms}. &1,010 &3,648 \\ \midrule
		
		Web Edu\cite{gleich04} & A web network. Source: \url{http://networkrepository.com/web-edu.php}. &3,031 &6,474 \\ \midrule
		
		Web EPA\cite{pajek11} & Network of web pages that link to \url{www.epa.gov}. Source: \url{http://networkrepository.com/web-EPA.php}. &4,271 &8,909\\ \midrule
		
		WikiTalk\cite{wikitalk} & Communication network among users of the Welsh Wikipedia. Source: \url{http://konect.cc/networks/wiki_talk_cy}. &2,101 & 3,951 \\ \midrule
		
		Yeast \cite{dongbo-2003-yeast} & A network of protein-protein interaction in budding yeast. Source: \url{http://vlado.fmf.uni-lj.si/pub/networks/data/bio/Yeast/Yeast.htm}. &2,284 &6,646 \\ \midrule
		
		Youtube\cite{soccompnets} & A Youtube friendship network. Source: \url{http://socialcomputing.asu.edu/pages/datasets}. &13,723 &76,764 \\
		
		\bottomrule
	\end{tabular}
\end{table*}

\subsection{Performance measures}

To test the performance of link prediction algorithms, we remove a set of links from a ground truth network and use it as a test set. A perfect link prediction algorithm should give higher scores to these removed links than links that initially did not exist in the ground truth network. We shall remove 10\% of the links in most experiments and use the remaining 90\% to train the algorithms. This procedure is per the standard practice in the link prediction literature. In experiments dealing with the robustness to sparsity, however, we will use different removal ratios to test the predictive power of the algorithms under various data scarcity conditions.

To quantify and assess the performance of the proposed link prediction approach, we use three standard performance measures in the field of link prediction: the area under the receiver operating curve (AUROC), the area under the precision-recall curve (AUPR), and top-precision (TPR). The AUROC can efficiently be computed as the probability that a removed link is assigned a higher score than an originally non-existing link. It has historically been the most used performance measure in the field \cite{Lu2011LinkPredictioninComplexNetworks}. However, since most networks are highly sparse, the AUROC may produce results close to the perfect score of 1, hiding in the process the poor quality of the predictions. This behavior is caused by the severe imbalance in size between the positive set (existing links) and the negative set (non-existing links), particularly in large networks \cite{yang2015EvaluatingLinkPredictionMethods,garcia2016limitations, Wang2016,muscoloni2017LocalRing, kerrache20}. 

Unlike the AUROC, the AUPR and TPR focus on the positive set, allowing them to avoid the pitfalls of the former \cite{davis06, yang2015EvaluatingLinkPredictionMethods}. The AUPR is computed by first constructing the precision-recall curve and then computing the area under it by numerical integration using either the simple trapezoidal rule or the more accurate nonlinear interpolation scheme proposed in \cite{davis06}.
TPR or top-$k$ precision \cite{yang2015EvaluatingLinkPredictionMethods} is defined as the ratio of removed links among the top $k$ scored links, where $k$ is  the total number of removed links \cite{Lu2011LinkPredictioninComplexNetworks,yang2015EvaluatingLinkPredictionMethods}. 
Due to its better reflection of the prediction quality and computational efficiency,  this paper follows the current practice of using TPR as the principal performance measure for evaluating link prediction algorithms \cite{Lu2011LinkPredictioninComplexNetworks,yang2015EvaluatingLinkPredictionMethods}. However, we also report the AUPR and AUROC results for completeness. 

To aggregate the performance results over networks with vastly different sizes and topological properties, we use the average significant ranking measure introduced in \cite{kerrache20}. This approach relies on statistical significance tests, which makes it more statistically robust than simply ranking the algorithms as in \cite{Wang2016, muscoloni2017LocalRing}. The average statistical rank of an algorithm is computed as follows. First, for each network, a two-tailed paired t-test is performed to compare the average performance of each pair of algorithms. If the statistical test is significant, the better algorithm is assigned the score of 1, whereas the other is assigned -1. If, on the other hand, the test is insignificant, both algorithms are assigned a score of 0. For each network, the algorithms are ranked based on these scores instead of the raw performance values. The ranks of an algorithm are then averaged over all networks to obtain its average significant rank. We follow the convention that the lower the rank, the better.

\subsection{Competing methods}
\begin{figure*}[!t]
	\centering
	\includegraphics[width=0.8\textwidth]{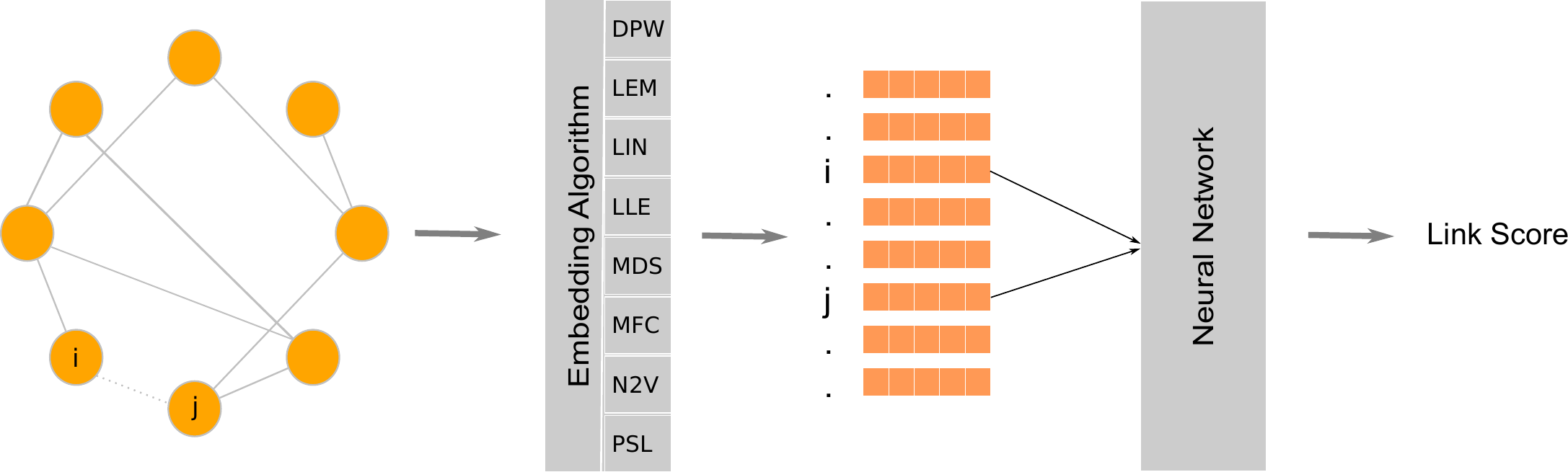}
	\caption{The graph-embedding-based link prediction framework used for performance evaluation.}
	\label{fig:framework}
\end{figure*}

We compare the proposed embedding method to the main state-of-the-art embedding methods, namely, DeepWalk (DPW) \cite{deepwalk}, Laplacian Eigenmaps (LEM) \cite{lapeigen}, Locally Linear Embedding (LLE) \cite{lle},  LINE: Large Information Networks Embedding (LIN) \cite{line}, Matrix Factorization (MFC), also known as Graph Factorization \cite{matfact09}, and Node2Vec (N2V) \cite{node2vec}. More details about these methods are given in Section \ref{sec:lr-embedding}. Graph embedding algorithms can be used to encode networks for various downstream tasks. Our focus in this work is on the link prediction task, and for that, we will use the embedding-based link prediction framework illustrated in Figure \ref{fig:framework}. First, the network is embedded using an embedding algorithm so that every node $i$ is assigned a code vector $x^i$. A neural network is then given inputs of the form $(x^i, x^j)$ that represents the edge $(i,j)$ and trained to discriminate between positive examples (existing edges) and negative examples (absent edges). The neural network can be trained on the whole set of couples or a sample of it, which may be necessary for large networks where the size of the negative class can be prohibitively large. 

\subsection{Implementation}
The proposed embedding algorithm is implemented in C++ using the functionalities of LinkPred, a high-performance library for link prediction \cite{linkpred}. The optimization tasks required in the embedding procedure are solved using the algorithm described in \cite{cgdescent-2006}, which is a conjugate gradient method with guaranteed descent. For the competing methods, we use the implementations provided by LinkPred. For the neural network, we use the feed-forward neural network implementation available in mlpack \cite{mlpack13, mlpack18}. We also use the data handling and performance evaluation functionalities of LinkPred in our performance evaluation procedure.

\subsection{Comparison of the different variants of the proposed method}
\label{sec:variants}
This experiment aims to compare various design decisions regarding the proposed method in terms of the choice of the objective function, popularity-similarity and local attraction combination approach, and reconstruction function. More precisely, we compare the performance of four variants described in Table \ref{tab:variants-desc} (in the table, PSL stands for Popularity-Similarity-Local attraction). We use the networks described in Table \ref{tab:data} and randomly remove 10\% of the existing edges and use them as a test set. We set the embedding dimension to 32 for all variants and the regularization coefficient $\lambda = 0.001$. For variants that use a neural network as a reconstruction function, we use an architecture with five layers having sizes 32, 16, 8, 4, and 2, respectively. We repeat the experiment 100 times for every network and report the average significant ranks on all networks based on TPR, AUPR, and AUROC in Figure \ref{fig:variants}. The detailed results are reported in the appendix in Section \ref{sec:app-variants}.

As Figure \ref{fig:variants} shows, reconstruction by neural network gives better results than by simple dot-product, albeit at a higher computational cost due to the learning process. Results also show that the PSL-NN-L2 variant, which uses the concatenation of popularity-similarity and local attraction embeddings with $L_2$ cost and a neural network as a reconstruction function, produces the best results under all three performance measures. Consequently, we shall henceforth use this variant as the default representative of the proposed method and refer to it as PSL for conciseness.
\begin{table*}[!t]
    \centering
    \caption{Description of the different variants of the proposed method.}
    \label{tab:variants-desc}
    \begin{tabular}{llll}
    \toprule
    Variant's name & Combination method & Objective function & Reconstruction function \\ \midrule
    PSL-NN-L1 & Concatenation & $L_1$ cost (Eq. \eqref{eq:ps-l1} and \eqref{eq:la-l1}) & Learned from data using a neural network\\ \midrule
    PSL-NN-L2 & Concatenation & $L_2$ cost (Eq. \eqref{eq:ps-l2} and \eqref{eq:la-l2}) & Learned from data using a neural network\\ \midrule
    PSL-NN-CO & Combined optimization & $L_2$ cost (Eq. \eqref{eq:psl-l2}) & Learned from data using a neural network\\ \midrule
    PSL-DP-L2 & Concatenation & $L_2$ cost (Eq. \eqref{eq:ps-l2} and \eqref{eq:la-l2}) &  Dot product\\
   \bottomrule
   \end{tabular}
\end{table*}

\begin{figure*}[!t]
	\centering
	\includegraphics[width=0.32\textwidth]{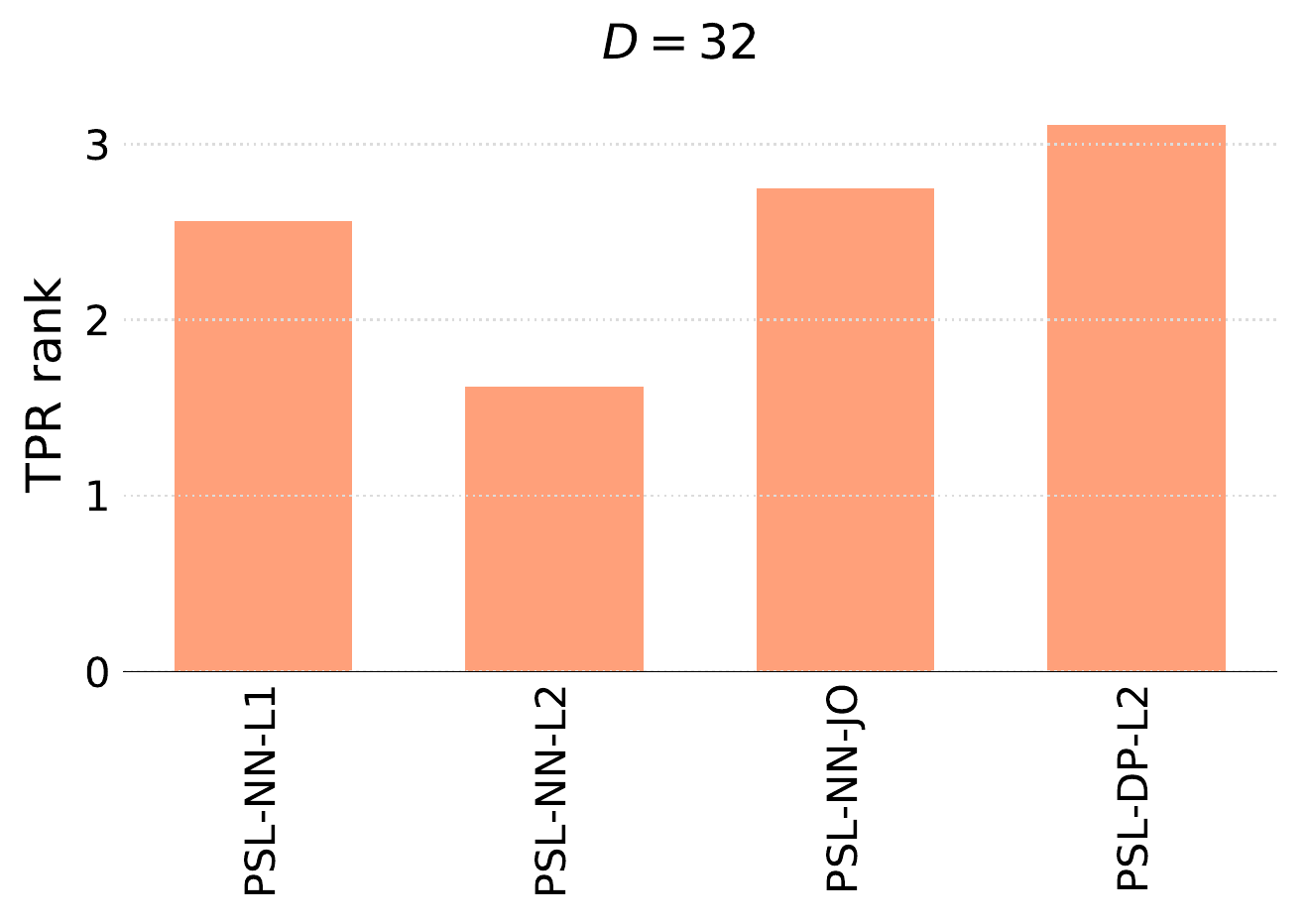}
	\includegraphics[width=0.32\textwidth]{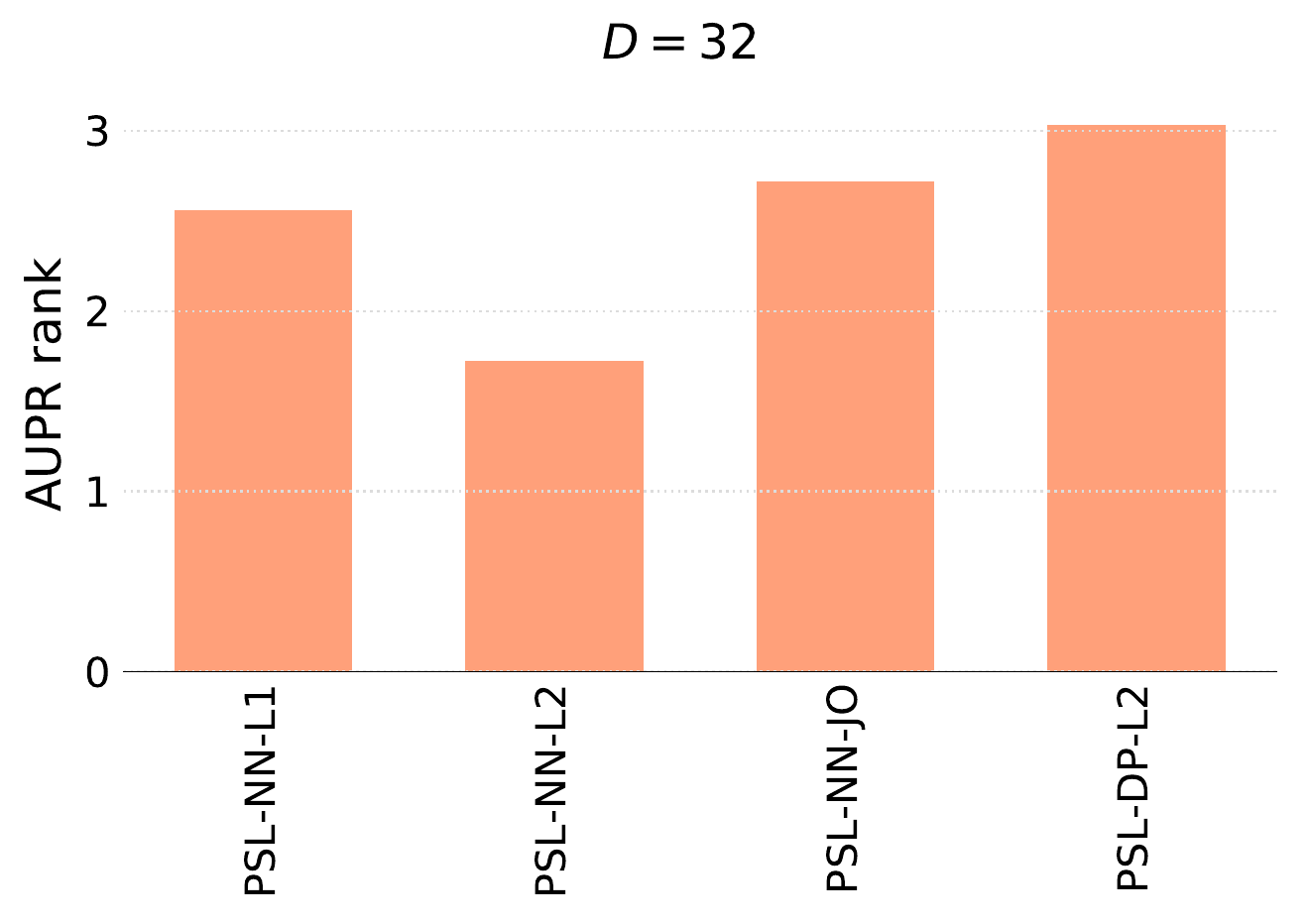}
	\includegraphics[width=0.32\textwidth]{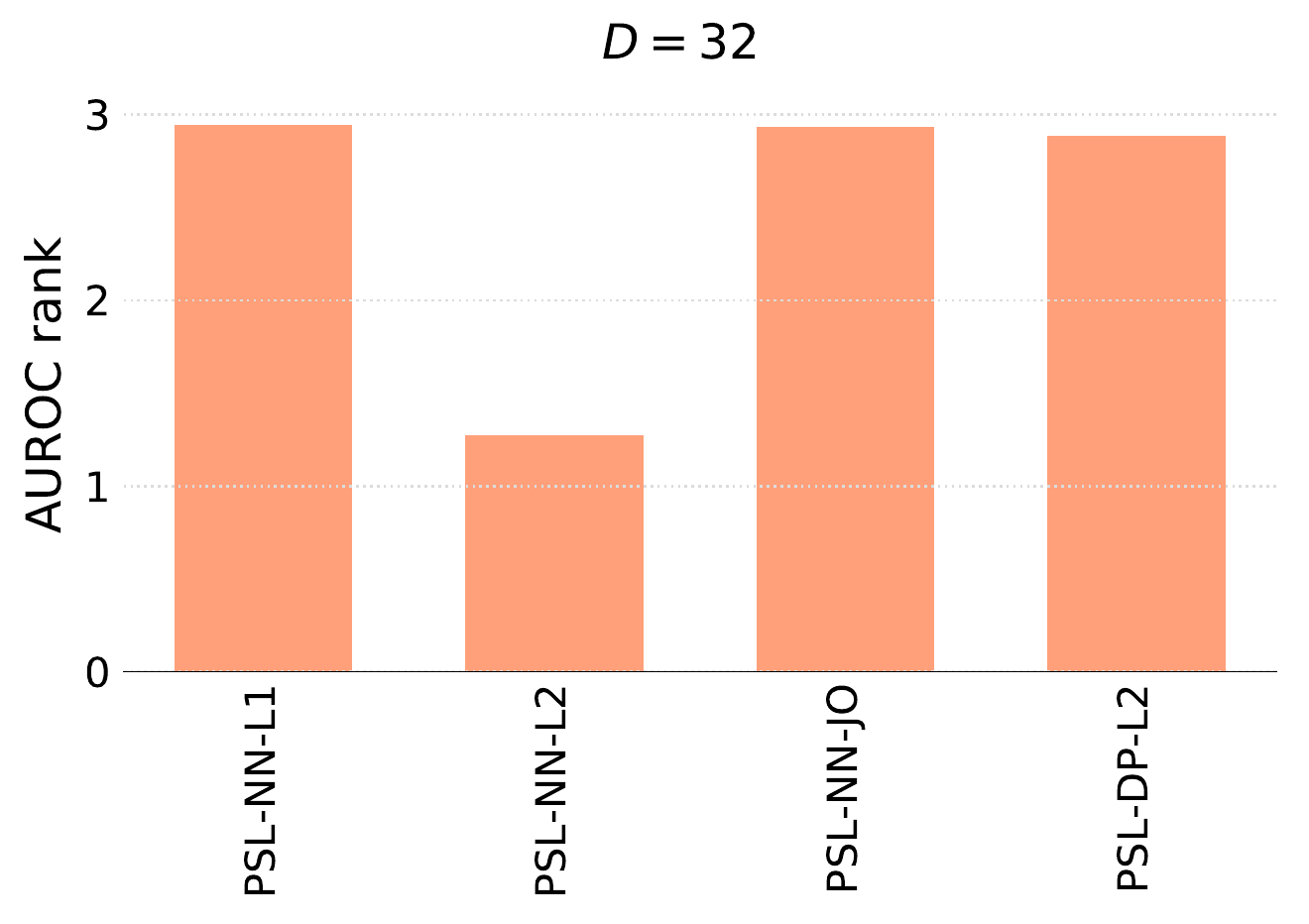}
	\caption{Average significant ranks of the different variants of the proposed method on networks with less than 1000 nodes based on TPR (left), AUPR (middle) and AUROC (right).}
	\label{fig:variants}
\end{figure*}

\subsection{Comparison against state-of-the-art embedding methods}
\label{sec:comp}
We compare the proposed method against two categories of methods. The first category consists of algebraic methods, namely, Laplacian Eigenmaps (LEM) \cite{lapeigen} and Locally Linear Embedding (LLE) \cite{lle}. These two methods reduce the embedding problem into an eigenvalue problem, hence the name algebraic. Due to the high computational cost of the resulting eigenvalue problems, the comparison against algebraic method is limited to relatively small networks. The second category contains DeepWalk (DPW) \cite{deepwalk}, LINE: Large Information Networks Embedding (LIN) \cite{line}, Matrix Factorization (MFC), also known as Graph Factorization \cite{matfact09}, Multi-Dimensional Scaling (MDS) \cite{mds}, and Node2Vec (N2V) \cite{node2vec}. MDS is a relatively old method compared to the other competing methods, but it is included in the experiment for completeness.

For algebraic methods, we use the networks of Table \ref{tab:data} where we randomly remove 10\% of the existing edges and use them as a test set. We set the embedding dimension to 8 for all algorithms and the regularization coefficient of PSL to $\lambda = 0.001$. All embedding methods are followed by a neural network for reconstruction. We use an architecture of five layers with sizes 32, 16, 8, 4, and 2, respectively. We repeat the experiment 100 times for every network and report the average significant ranks over all networks based on TPR, AUPR, and AUROC in Figure \ref{fig:alg}. The detailed results are reported in the appendix in Section \ref{sec:app-comp}, Table \ref{tab:comp-alg}. The results show that the proposed method outperforms Laplacian Eigenmaps (LEM) and Locally Linear Embedding (LLE) under all three performance measures, TRP, AUPR and AUROC.
\begin{figure*}[!t]
	\centering
	\includegraphics[width=0.32\textwidth]{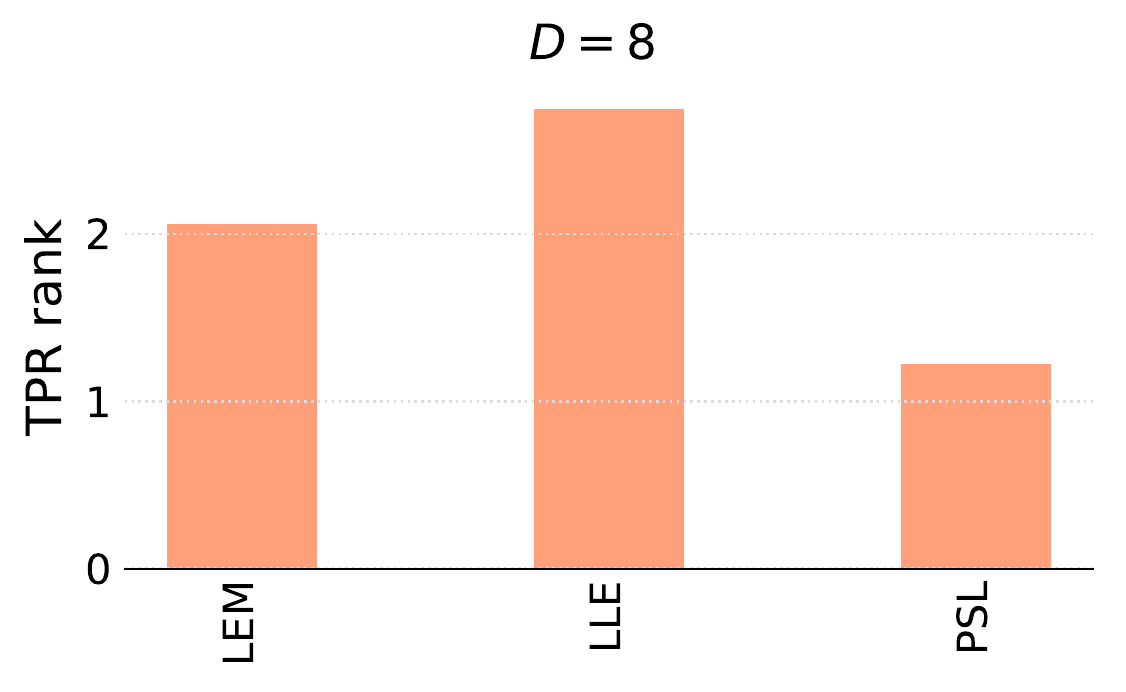}
	\includegraphics[width=0.32\textwidth]{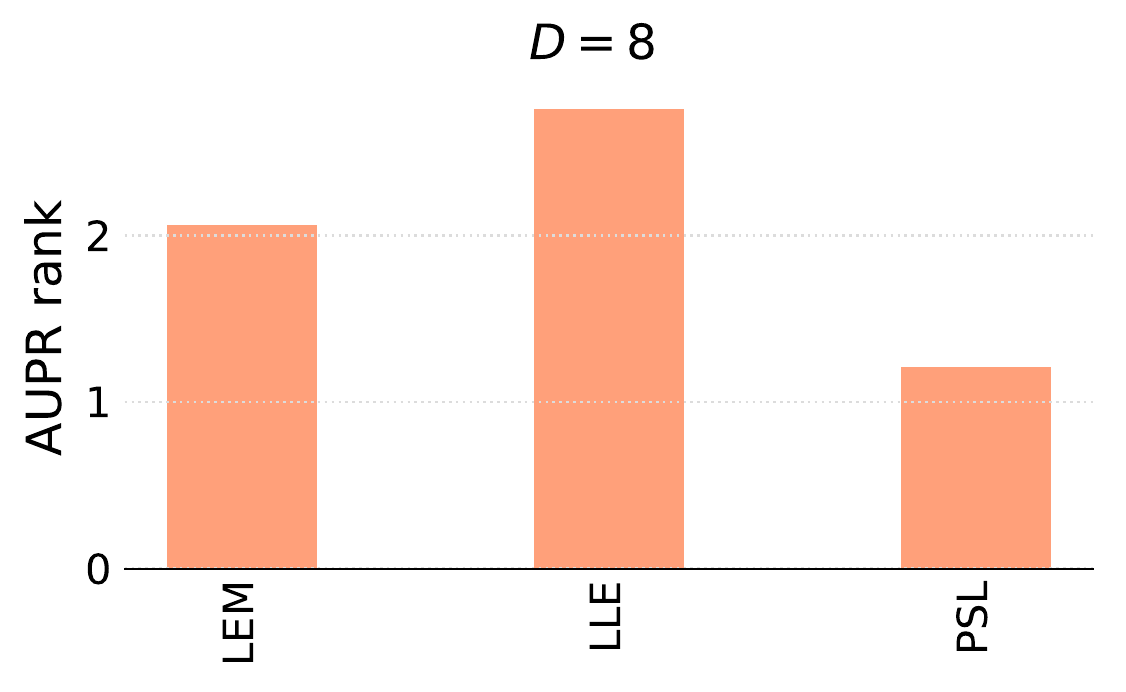}
	\includegraphics[width=0.32\textwidth]{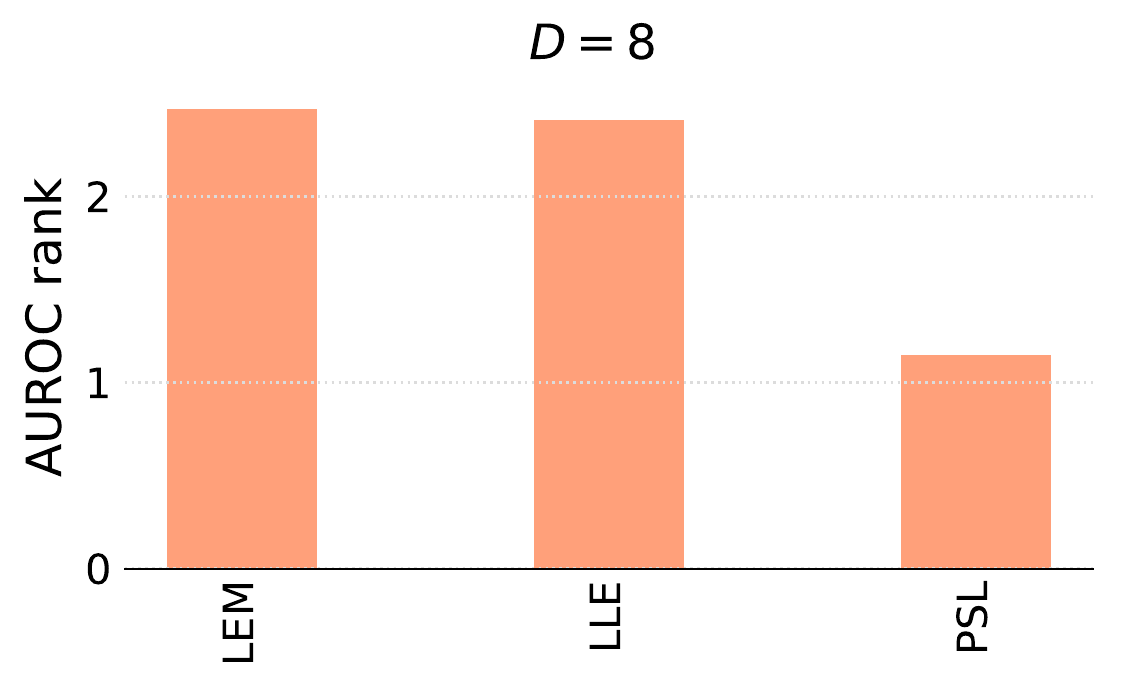}
	\caption{Average significant ranks comparison against algebraic embedding methods on networks with less than 1000 nodes based on TPR (left), AUPR (middle) and AUROC (right).}
	\label{fig:alg}
\end{figure*}

For non-algebraic methods, we run three sets of experiments. The first two are conducted on the networks of Table \ref{tab:data}, whereas the third is conducted on the larger networks described in Table \ref{tab:data-large}. For the networks of Table \ref{tab:data}, we set the embedding dimension to eight in the first experiment and 32 for the second one. We set the regularization coefficient of PSL to $\lambda = 0.001$. For reconstruction, we add a neural network after all embedding algorithms. The network comprises five layers with sizes 32, 16, 8, 4, and 2, respectively. We repeat the experiment 100 times for every network and report the average significant ranks over all networks based on TPR, AUPR, and AUROC in Figure \ref{fig:nalg-small}. The detailed results are included in the appendix in Section \ref{sec:app-comp}, Table \ref{tab:nalg-small-tpr}, \ref{tab:nalg-small-aupr}, and \ref{tab:nalg-small-auroc}. 
As shown in Figure \ref{fig:nalg-small}, the proposed method (PSL) gives the best results under all three performance measures followed by Matrix Factorization (MFC) when the embedding dimension is eight and DeepWalk (DPW) when the dimension is set to 32. 

\begin{figure*}[!t]
	\centering
	\includegraphics[width=0.32\textwidth]{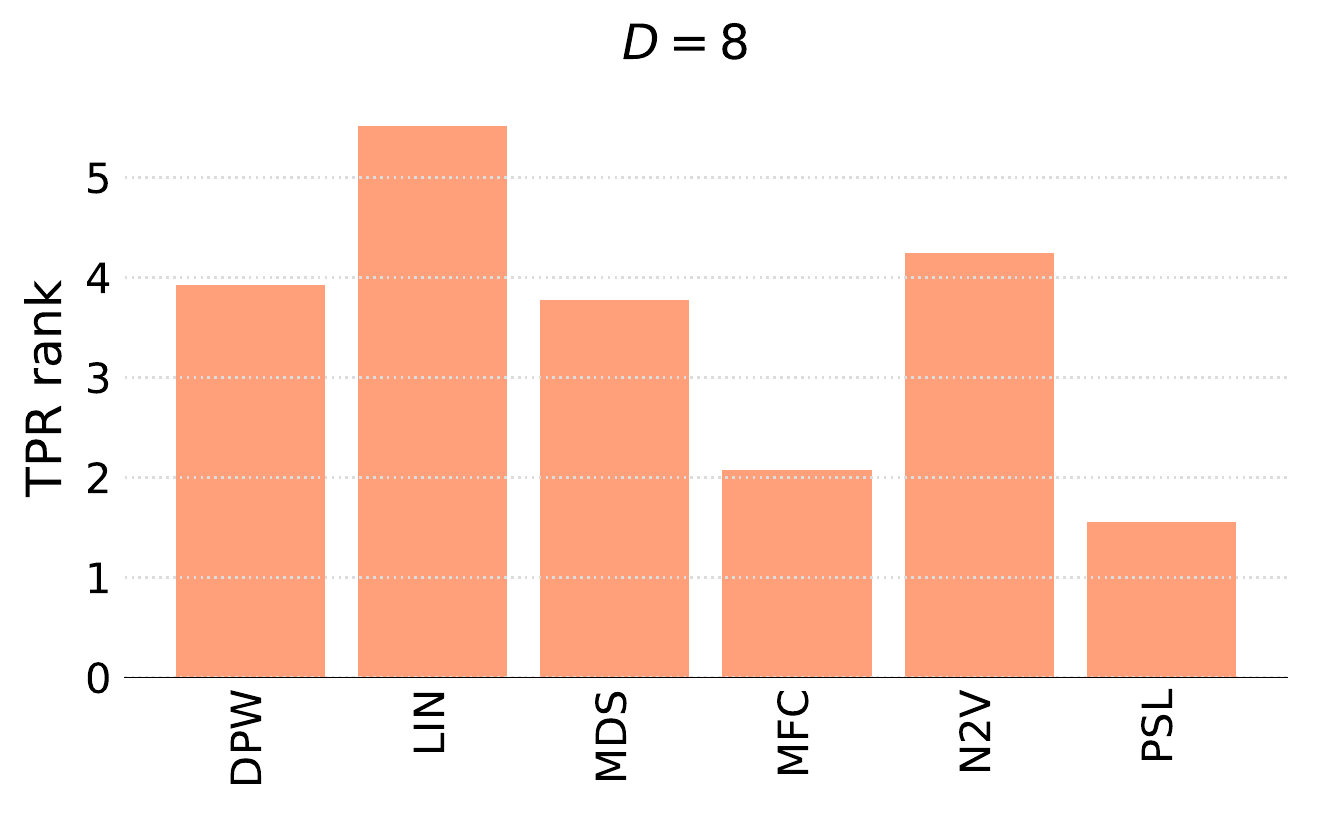}
	\includegraphics[width=0.32\textwidth]{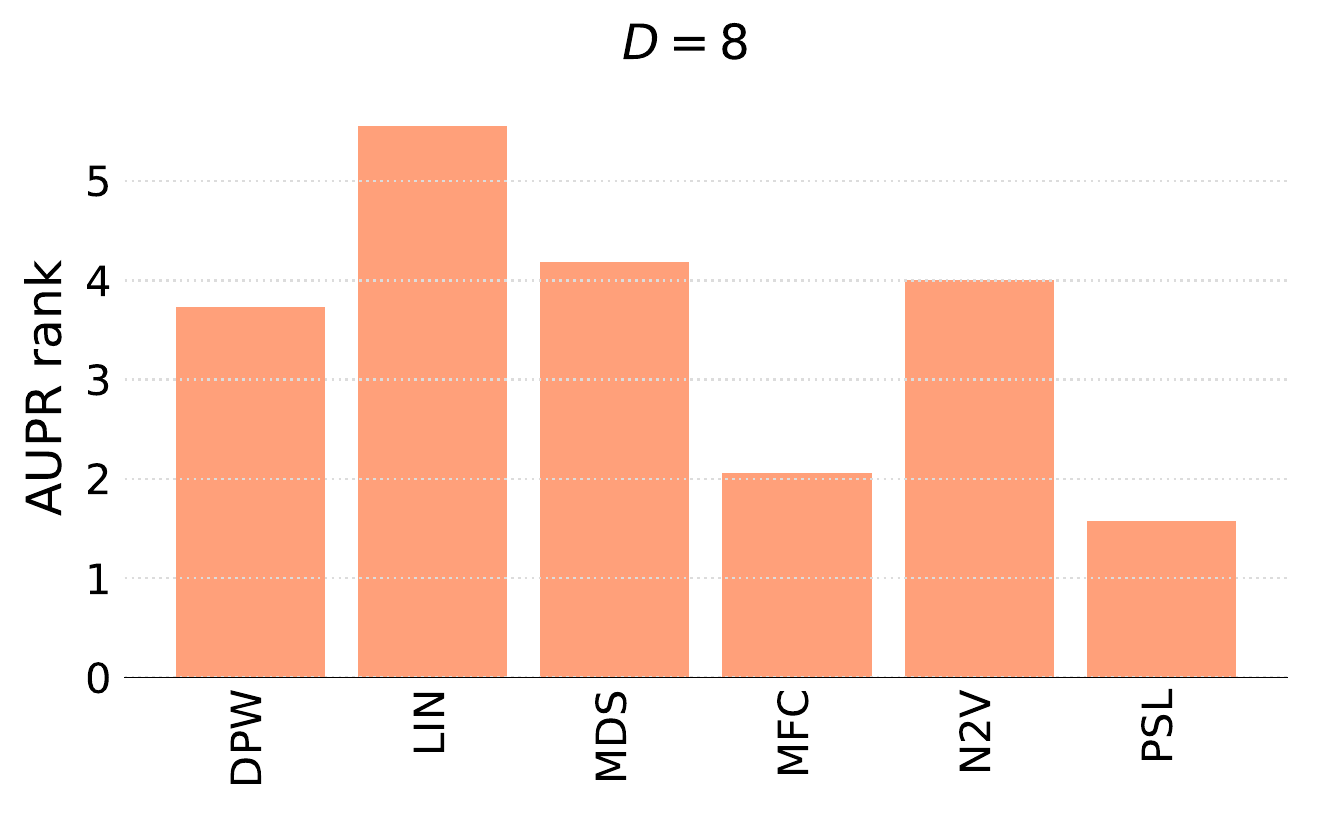}
	\includegraphics[width=0.32\textwidth]{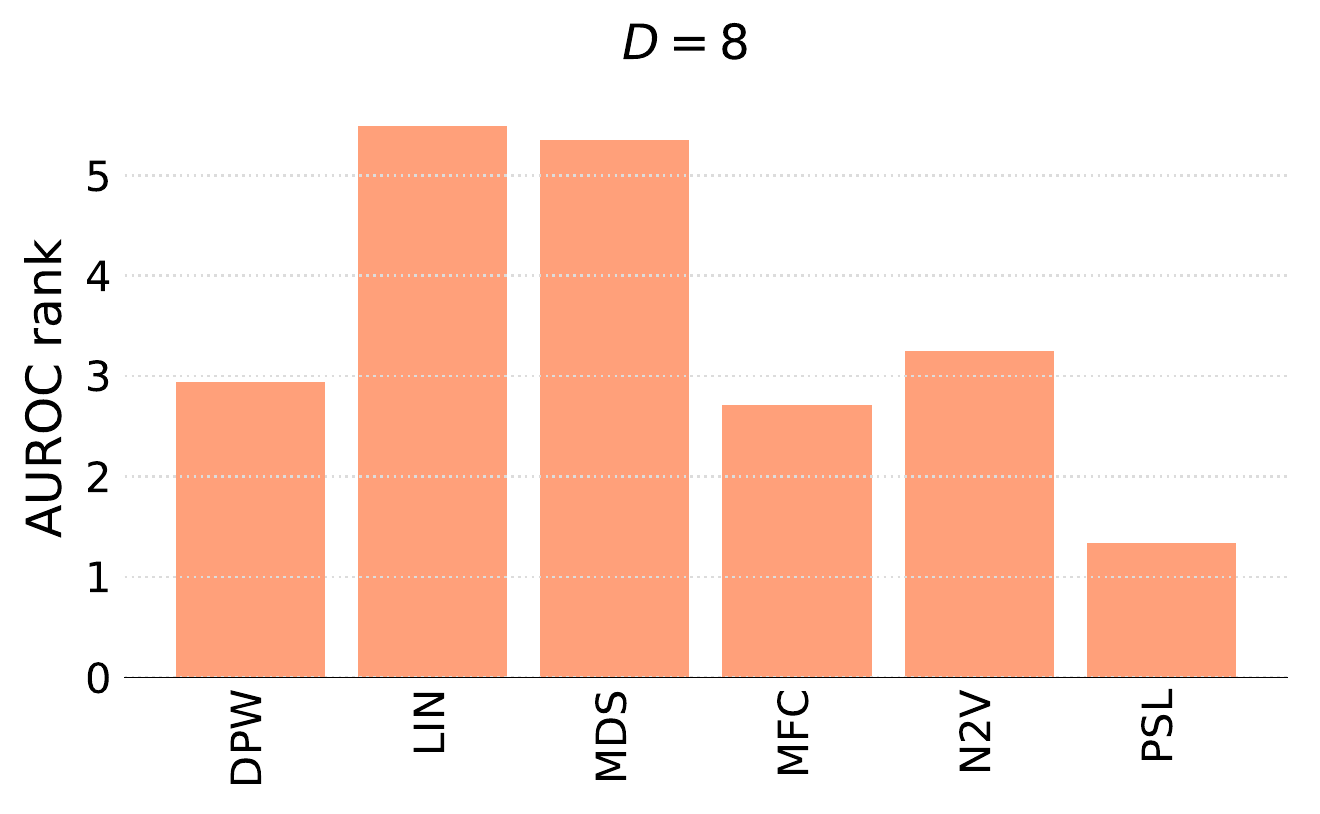}
	
	\includegraphics[width=0.32\textwidth]{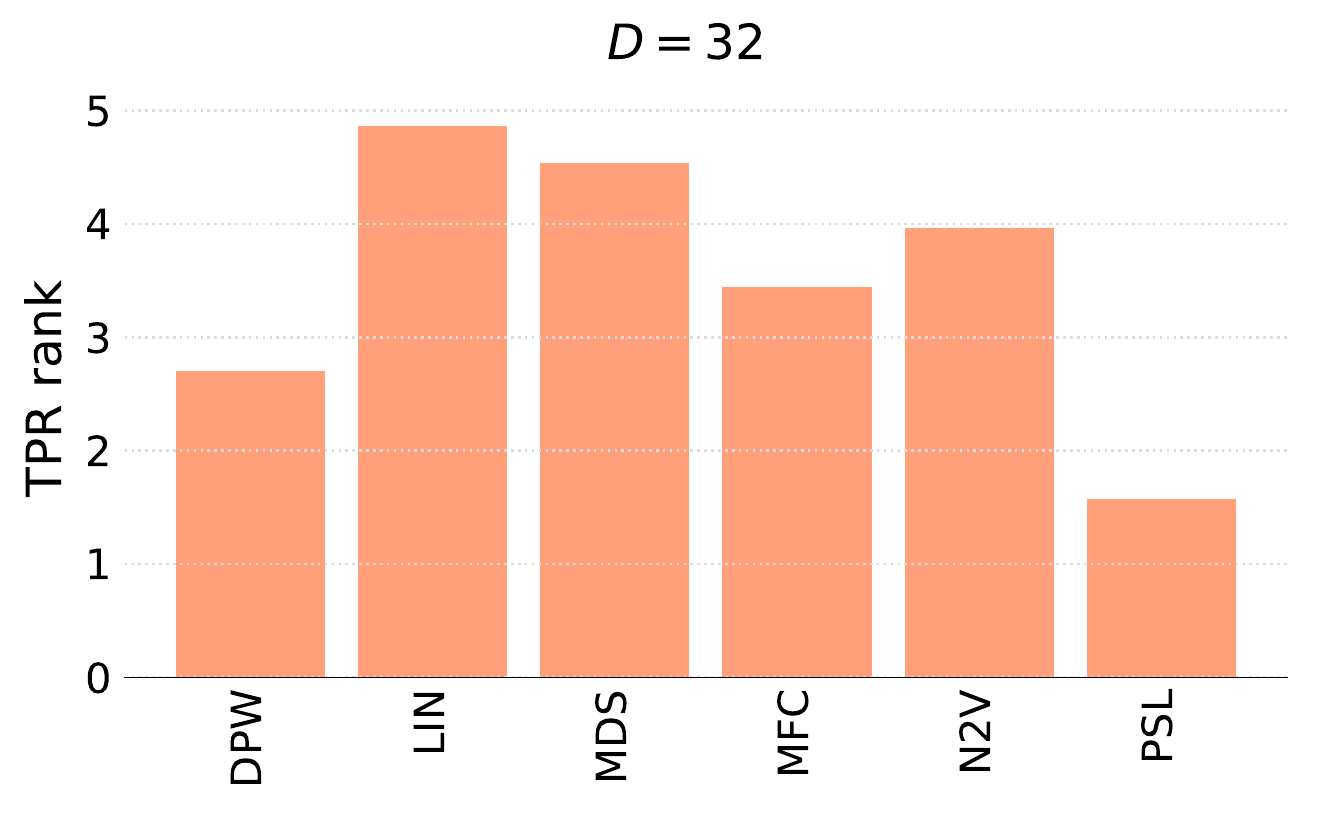}
	\includegraphics[width=0.32\textwidth]{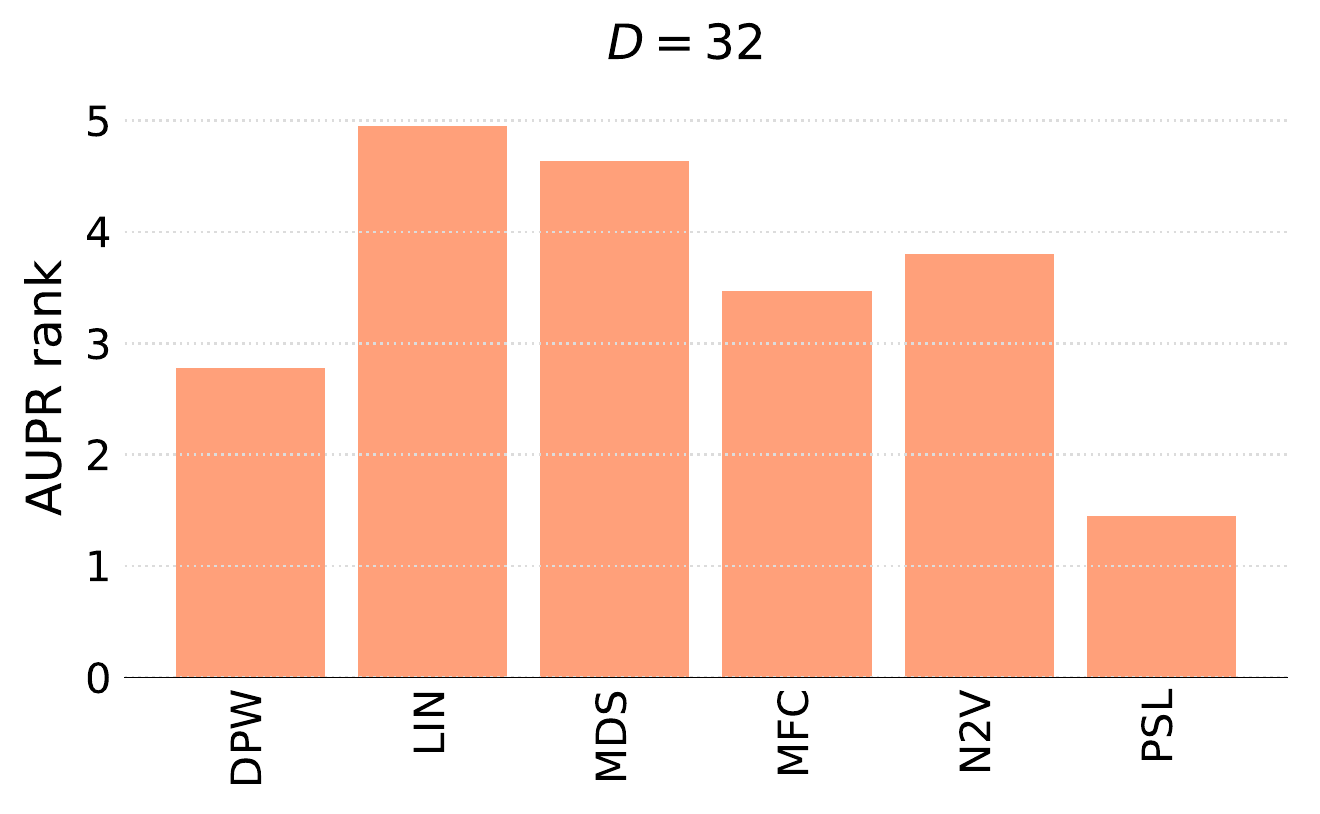}
	\includegraphics[width=0.32\textwidth]{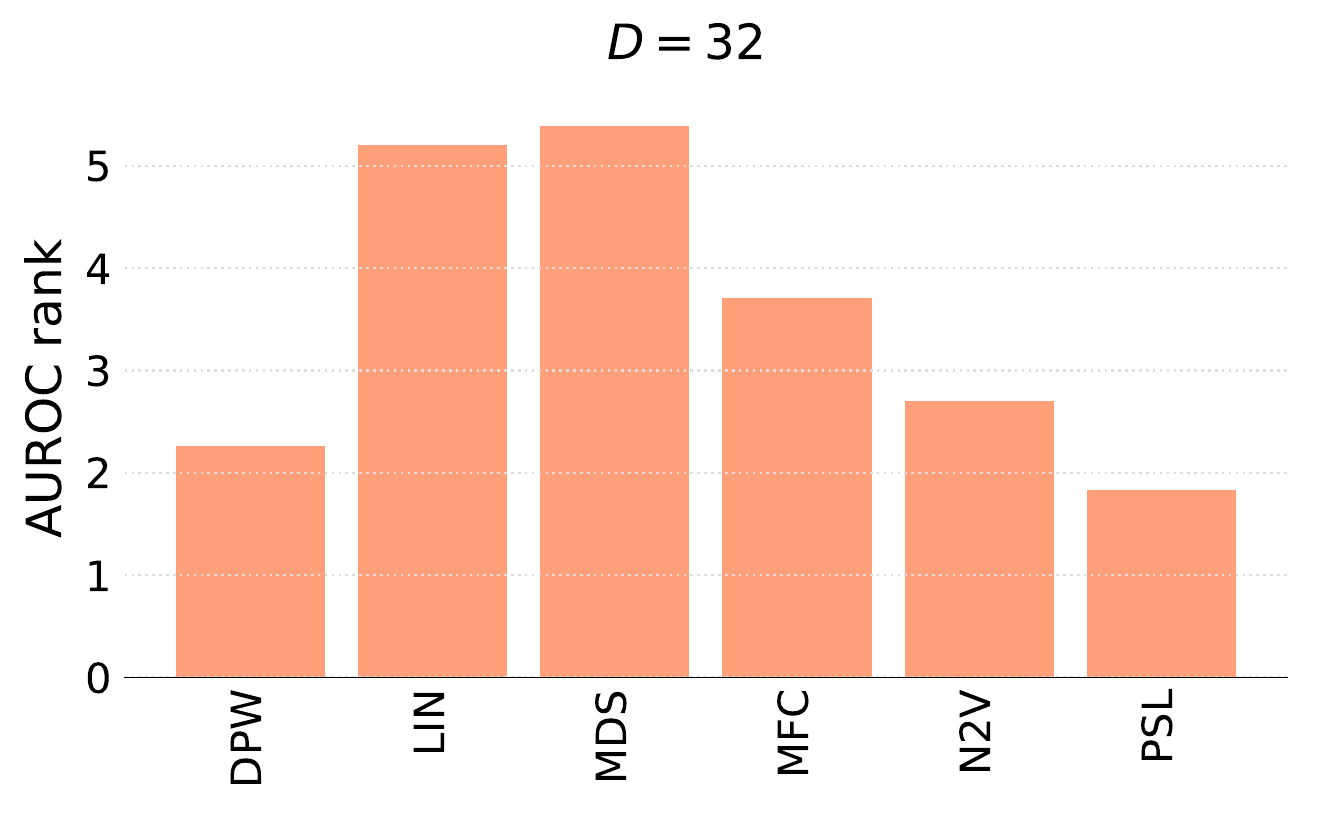}
	
	\caption{Average significant ranks comparison against non-algebraic methods with embedding dimensions 8 and 32 on networks with less than 1000 nodes based on TPR (left), AUPR (middle) and AUROC (right).}
	\label{fig:nalg-small}
\end{figure*}

\begin{figure*}[!t]
	\centering
	\includegraphics[width=0.32\textwidth]{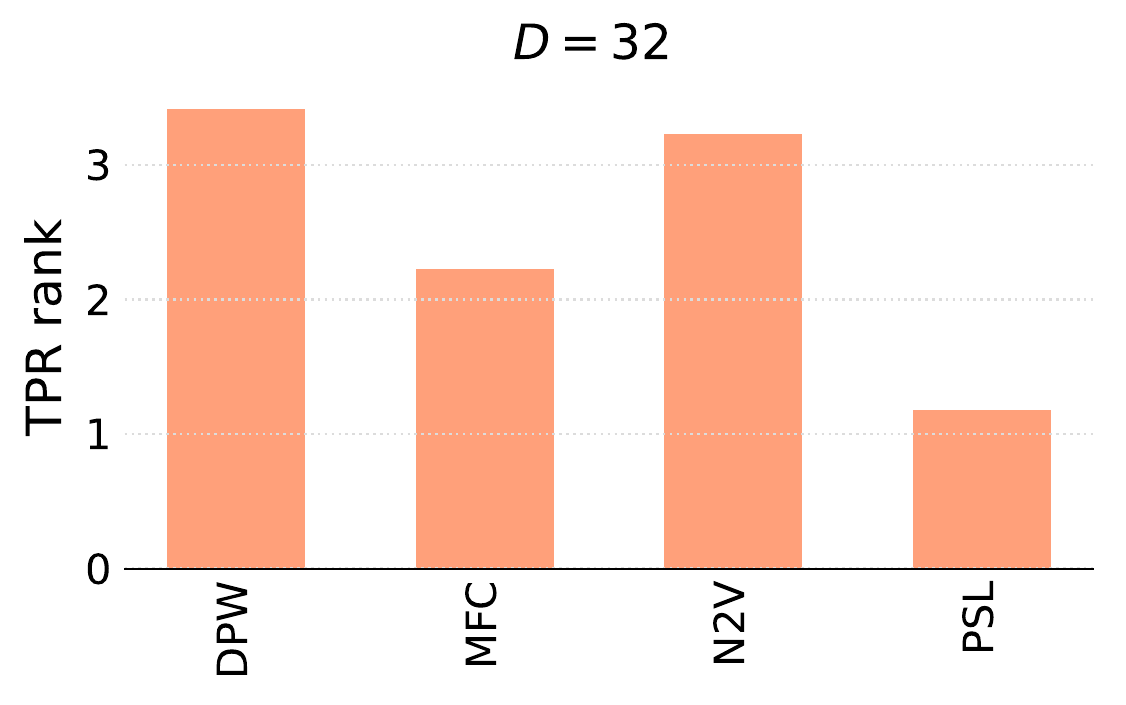}
	\includegraphics[width=0.32\textwidth]{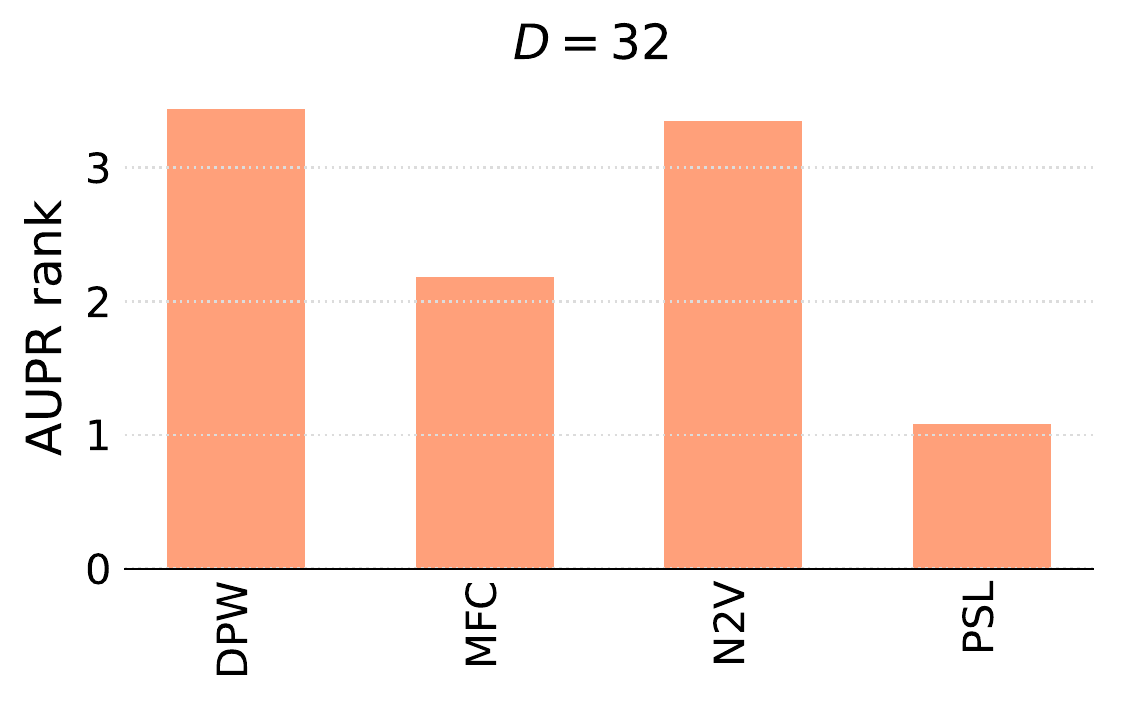}
	\includegraphics[width=0.32\textwidth]{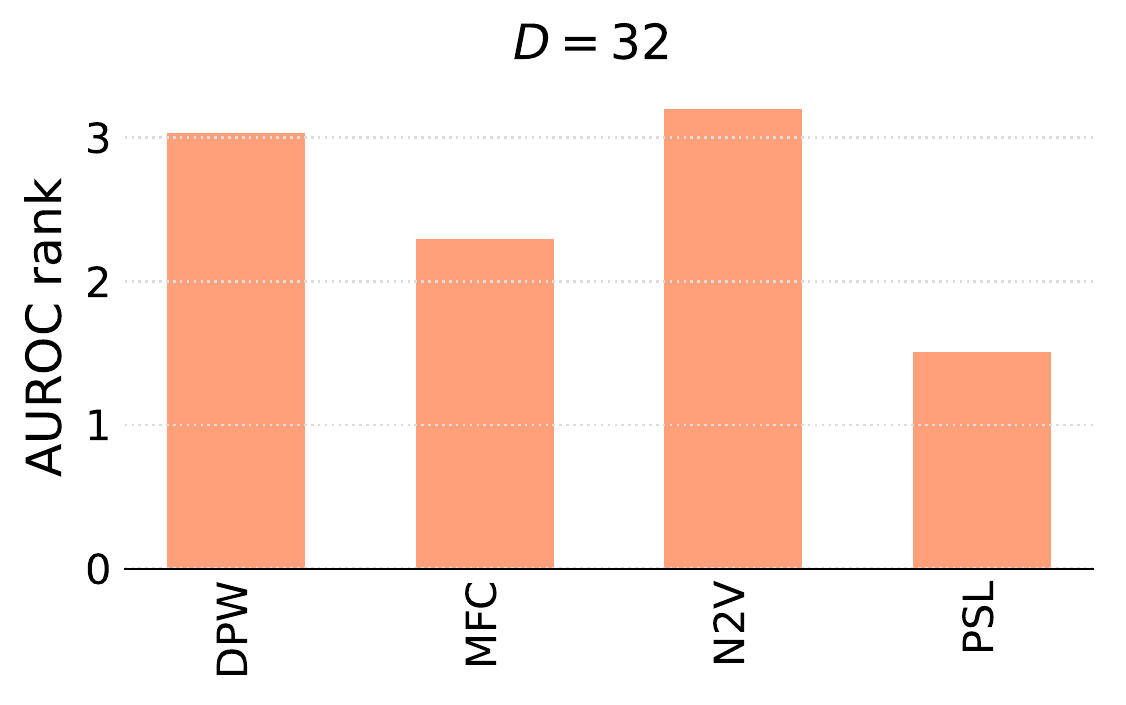}
	\caption{Average significant ranks comparison on networks with more than 1000 nodes based on TPR (left), AUPR (middle) and AUROC (right).}
	\label{fig:large}
\end{figure*}

We conduct the third experiment on the larger networks of Table \ref{tab:data-large}. To keep the computational requirements reasonable, we adjust the experimental settings as follows. First, we compare only the top four methods from the previous experiment: DPW, MFC, N2V, and PSL. We use a logistic regression model as a classifier instead of the neural network and use all positive links and only 0.1\% of the negative links to train it. Similarly, we use all positive links and 0.1\% of negative links to compute the performance measures. We limit the number of iterations in the two optimization tasks of PSL to 100 each. We set the embedding dimension to 32 and run ten trials per network. In each trial, we randomly remove 10\% of the exiting edges and use them as a test set. Figure \ref{fig:large} shows the obtained average significant ranks based on TPR, AUPR, and AUROC. The detailed per-network results are reported  in the appendix in Section \ref{sec:app-comp}, Table \ref{tab:large}. The results show that the proposed approach outperforms the competing methods in larger networks with an even more considerable margin than in small networks, especially in TPR and AUPR. MFC remains in the second position, but unlike in small networks, N2V performs slightly better than DPW.

\subsection{Effect of the embedding dimension}
\begin{figure*}[!t]
	\centering
	\includegraphics[width=\textwidth]{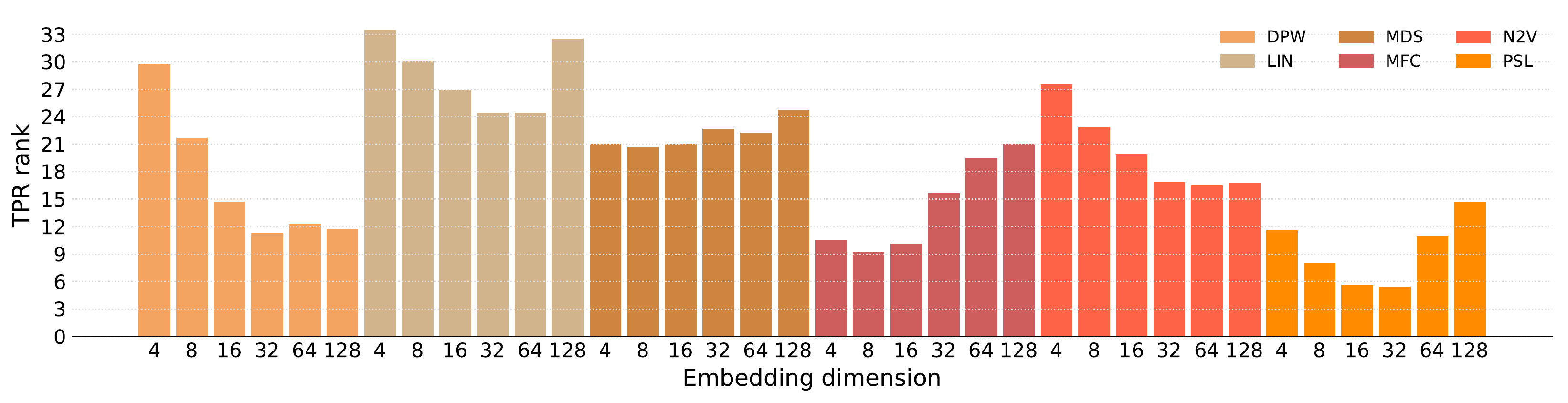}

	\includegraphics[width=\textwidth]{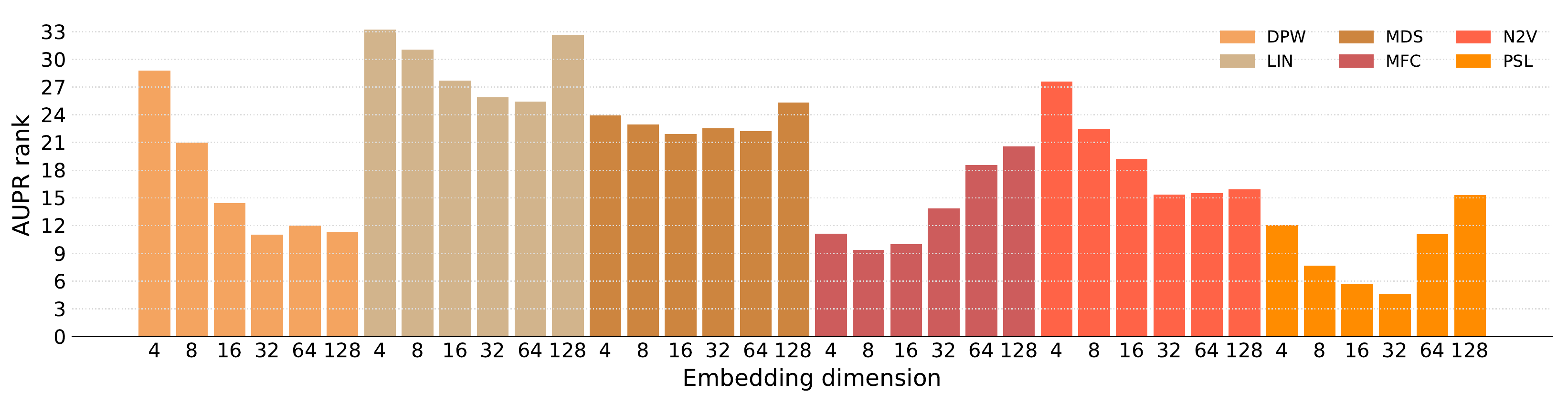}

	\includegraphics[width=\textwidth]{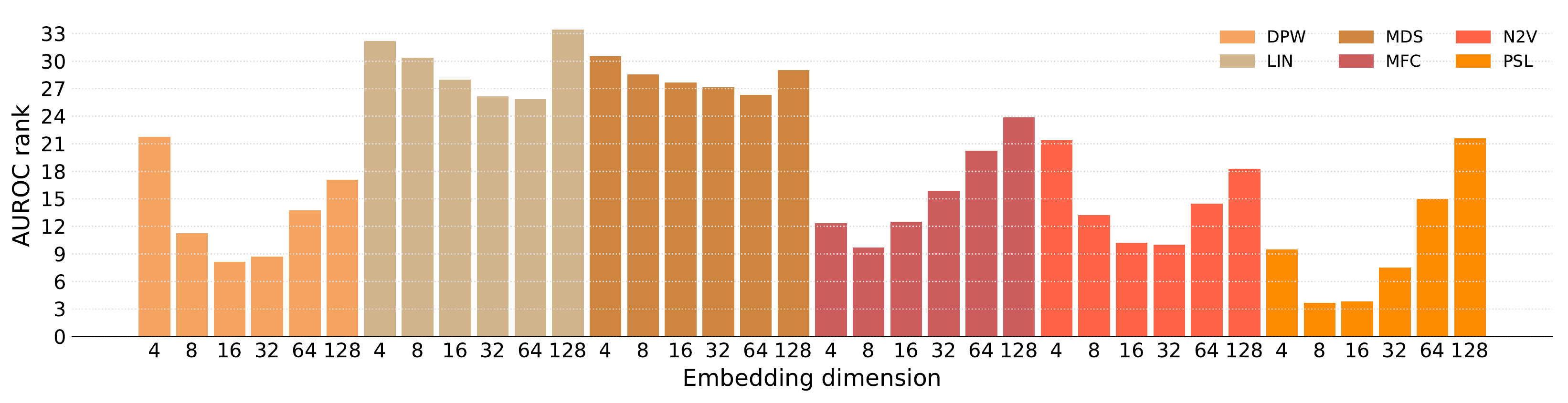}
	\caption{Average significant ranks on networks with less than 1000 nodes using different embedding dimensions. The ranks are based on the TPR (top), AUPR (middle), and AUROC (bottom).}
	\label{fig:dim}
\end{figure*}

To test the effect of the embedding dimension on the algorithm's performance, we compare the link prediction results obtained by our algorithm to the other embedding methods using different embedding dimensions. We vary the embedding dimension for each embedding method and train a feed-forward neural network to discriminate between connected and disconnected couples. We run the experiment on the networks shown in Table \ref{tab:data}. For each network, we run a hundred test runs, randomly removing 10\% of the links and using them as a test set. We report the average significant rank based on TPR, AUPR, and AUROC in Figure \ref{fig:dim}. We can observe that the proposed method produces the best results under all three performance measures and for several embedding dimensions. Of course, a good choice of the dimensionality of the embedding can be obtained using model selection by validation. However, this experiment shows that the algorithm is robust to this choice. For instance, Figure \ref{fig:dim} shows that any choice among 8, 16, and 32 produces the best overall results under all three performance measures.

\subsection{Evaluating the robustness of the proposed algorithm}
\begin{figure*}[!t]
	\centering
	\includegraphics[width=0.32\textwidth]{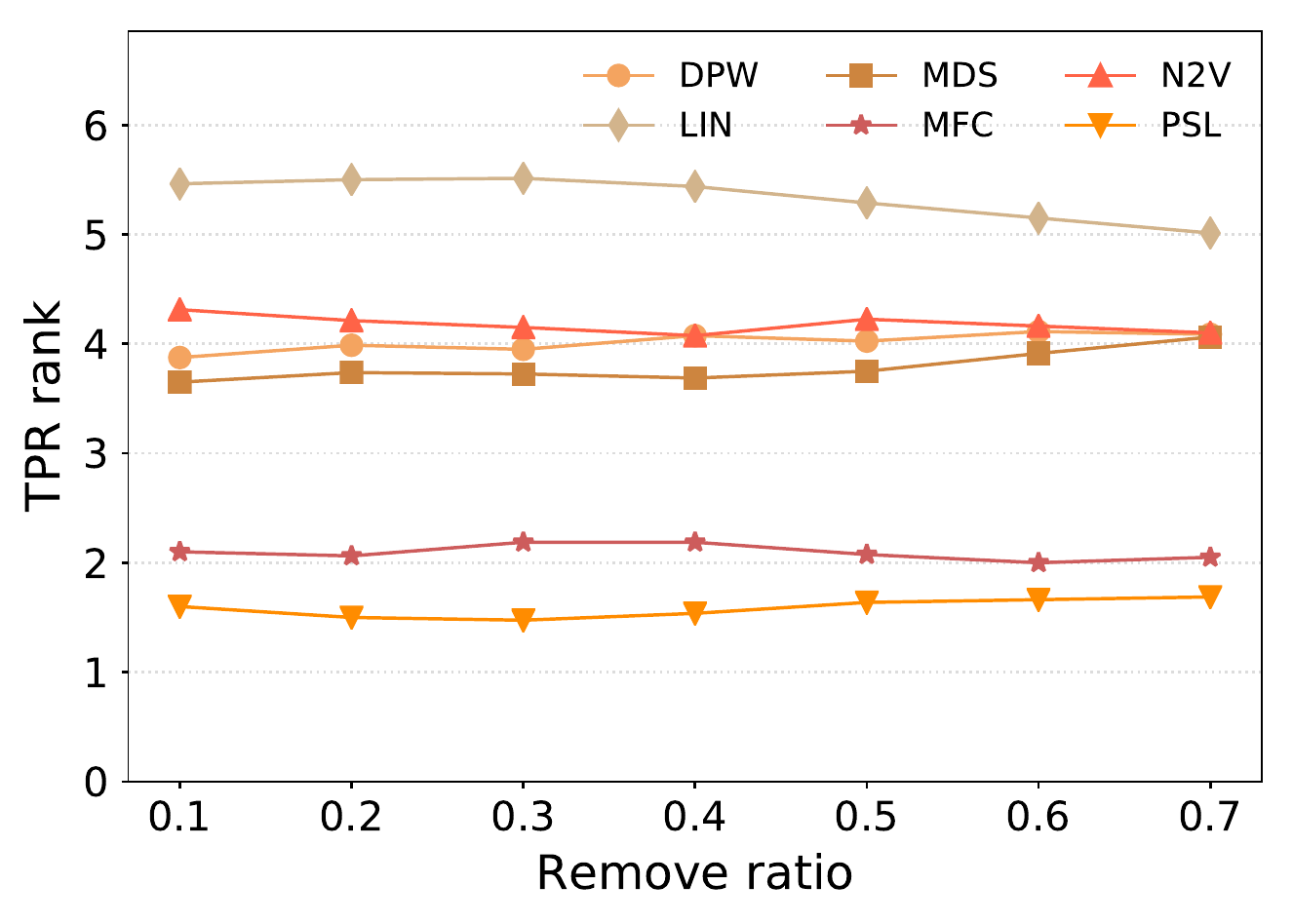}
    \includegraphics[width=0.32\textwidth]{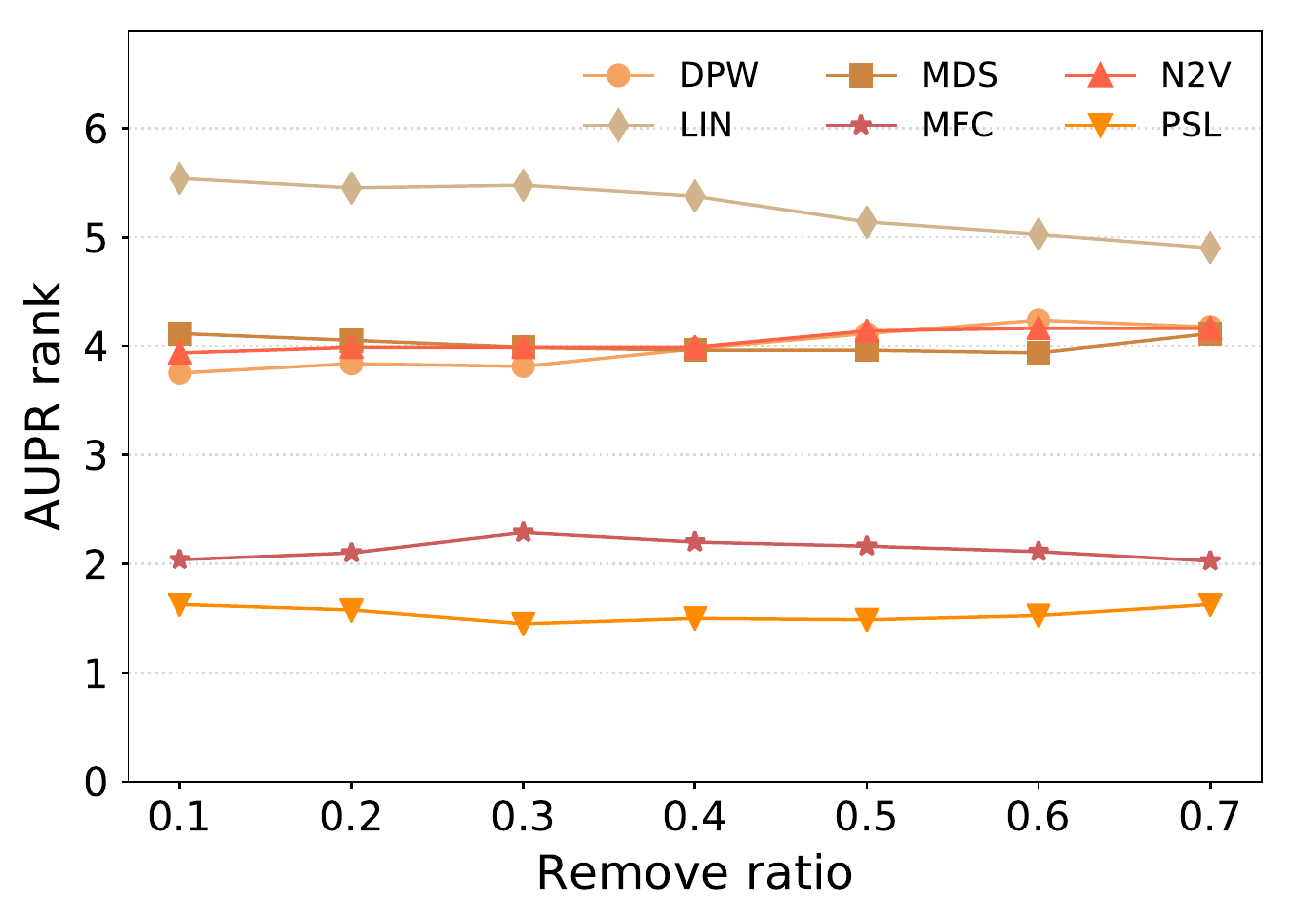}
    \includegraphics[width=0.32\textwidth]{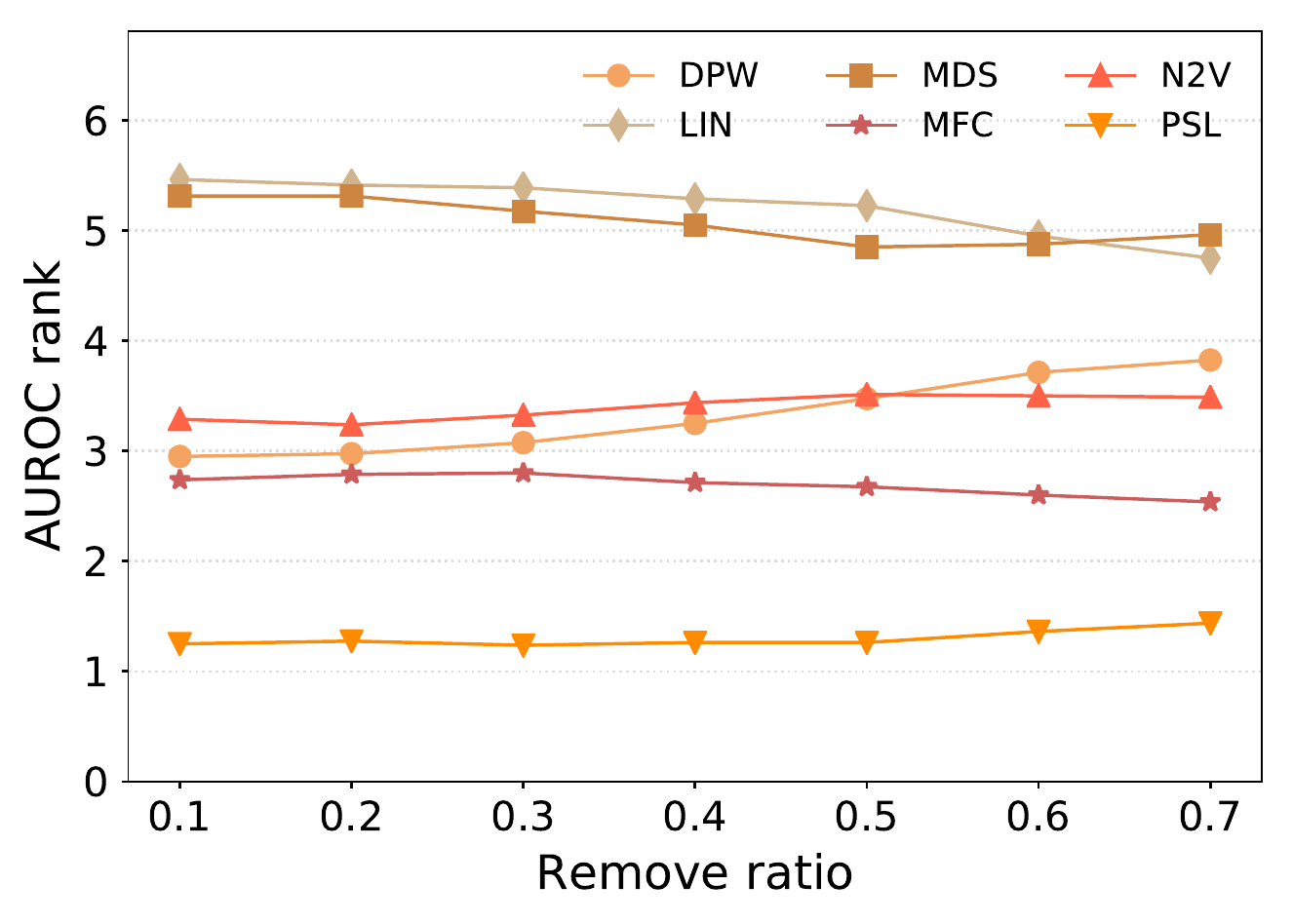}
	\caption{Average significant ranks comparison against non-algebraic methods with different edge remove ratios on networks with less than 1000 nodes. The ranks are based on TPR (left), AUPR (middle), and AUROC (right).}
	\label{fig:ratio}
\end{figure*}

To test the robustness of the proposed algorithm, we increase the ratio of removed links from 10\% to 70\% with an increment of 10\% and compare its performance to competing methods. We report the average significant ranks for the removal ratios 10\% and 70\% and the rank obtained over all ratios. For each network and removal ratio, we run 100 test runs. A feed-forward neural network is used as a classifier. As shown in Figure \ref{fig:ratio}, the proposed method maintains its superiority at all remove ratios and under all performance measures, proving its robustness even when a large proportion of the network topology is hidden.

\section{Conclusion}
\label{sec:conclusion}
This paper introduced a new graph embedding approach that combines the popularity-similarity and local attraction paradigms. The embedding problem is split and reduced to two model fitting problems solved via local optimization. The embedding algorithm is used for link prediction by feeding its output to a neural network trained to discriminate between connected and disconnected couples. 

Extensive experiments demonstrate that the proposed method outperforms state-of-the-art embedding algorithms such as DeepWalk, Laplacian Eigenmaps, Locally Linear Embedding,  LINE: Large Information Networks Embedding, and Node2Vec. The results also show that the algorithm is resilient to data scarcity and robust to the choice of the embedding dimension.

In future work, we propose to investigate the effectiveness of the proposed embedding approach for other downstream tasks such as the long-term evolution of temporal networks, community detection, and information spreading prediction. Implementation-wise, we plan to implement and test faster implementations based on stochastic gradient descent-type algorithms on GPUs to be able to handle very large networks. Developing better sub-sampling strategies that reduce the computation cost while preserving enough topological information to obtain a faithful embedding is also an exciting research direction.


%

\appendices

\section{Further results}
This appendix contains detailed results of the various experiments conducted in the performance evaluation section of the paper.

\subsection{Detailed results for the experiment of Section \ref{sec:variants}}
\label{sec:app-variants}
Table \ref{tab:variants} contains the per-network TPR, AUPR, and AUROC results of the different variants of the proposed method as described in Section \ref{sec:variants}.
\begin{table*}[!t]
    \centering
    \caption{Performance results obtained by the different variants of the proposed embedding method on networks with less than 1000 nodes. For each, performance measure, the best results significant with $p$-value 0.05 are shown in bold. The last row shows the average significant rank of each variant; lower ranks are better.}
    \label{tab:variants}
\begin{tabular}{lcccccccccccc}
	\toprule
	                            &             \multicolumn{4}{c}{TPR}              &             \multicolumn{4}{c}{AUPR}             &            \multicolumn{4}{c}{AUROC}             \\ 
	                            \cmidrule(lr){2-5} \cmidrule(lr){6-9} \cmidrule(lr){10-13}
	Network                     &   \rotatebox{270}{PSL-NN-L1}    &   \rotatebox{270}{PSL-NN-L2}    &   \rotatebox{270}{PSL-NN-CO}    &   \rotatebox{270}{PSL-DP-L2}    &   \rotatebox{270}{PSL-NN-L1}    &   \rotatebox{270}{PSL-NN-L2}    &   \rotatebox{270}{PSL-NN-CO}    &   \rotatebox{270}{PSL-DP-L2}    &   \rotatebox{270}{PSL-NN-L1}    &   \rotatebox{270}{PSL-NN-L2}    &   \rotatebox{270}{PSL-NN-CO}    &   \rotatebox{270}{PSL-DP-L2}    \\ \midrule
	ACM2009 Contacts            & \textbf{0.237} &     0.229      &     0.193      &     0.228      & \textbf{0.190} &     0.175      &     0.140      &     0.179      &     0.759      & \textbf{0.764} &     0.719      & \textbf{0.767} \\ \midrule
	C.Elegans Metabolic         &     0.141      &     0.165      & \textbf{0.180} &     0.122      &     0.066      &     0.082      & \textbf{0.093} &     0.053      &     0.828      & \textbf{0.875} &     0.832      & \textbf{0.880} \\ \midrule
	C.Elegans Neural            &     0.088      & \textbf{0.119} &     0.109      &     0.099      &     0.040      & \textbf{0.065} &     0.055      &     0.056      &     0.815      & \textbf{0.881} &     0.866      &     0.874      \\ \midrule
	CPAN Authors                &     0.096      &     0.103      & \textbf{0.106} &     0.100      & \textbf{0.035} & \textbf{0.037} & \textbf{0.037} &     0.030      &     0.872      &     0.897      & \textbf{0.917} &     0.692      \\ \midrule
	Centrality Literature       &     0.208      & \textbf{0.225} &     0.123      &     0.154      &     0.130      & \textbf{0.140} &     0.067      &     0.085      &     0.838      & \textbf{0.878} &     0.806      &     0.854      \\ \midrule
	Chesapeake Lower            & \textbf{0.269} & \textbf{0.291} & \textbf{0.274} &     0.150      &     0.217      & \textbf{0.235} &     0.211      &     0.117      &     0.856      & \textbf{0.879} &     0.839      &     0.763      \\ \midrule
	Chesapeake Middle           &     0.346      & \textbf{0.370} &     0.332      &     0.275      &     0.323      & \textbf{0.343} &     0.305      &     0.264      &     0.869      & \textbf{0.885} &     0.857      &     0.819      \\ \midrule
	Chesapeake Upper            & \textbf{0.344} & \textbf{0.354} &     0.263      &     0.229      & \textbf{0.280} & \textbf{0.291} &     0.204      &     0.170      &     0.847      & \textbf{0.866} &     0.799      &     0.790      \\ \midrule
	Codeminer                   &     0.002      & \textbf{0.029} &     0.005      & \textbf{0.028} &     0.001      & \textbf{0.006} &     0.002      & \textbf{0.007} & \textbf{0.657} & \textbf{0.667} &     0.640      &     0.574      \\ \midrule
	Cypress Dry                 &     0.315      & \textbf{0.340} &     0.261      &     0.158      & \textbf{0.273} & \textbf{0.278} &     0.204      &     0.130      &     0.881      & \textbf{0.916} &     0.868      &     0.825      \\ \midrule
	Cypress Wet                 & \textbf{0.353} & \textbf{0.366} &     0.264      &     0.195      & \textbf{0.309} & \textbf{0.302} &     0.199      &     0.136      &     0.886      & \textbf{0.916} &     0.862      &     0.830      \\ \midrule
	DNA Citation                & \textbf{0.018} & \textbf{0.023} & \textbf{0.035} & \textbf{0.033} &     0.023      &     0.020      & \textbf{0.030} & \textbf{0.029} &     0.610      &     0.619      & \textbf{0.653} &     0.639      \\ \midrule
	DNA Citation CC             & \textbf{0.047} & \textbf{0.032} & \textbf{0.027} & \textbf{0.033} & \textbf{0.043} &     0.026      &     0.030      &     0.031      & \textbf{0.600} & \textbf{0.626} & \textbf{0.608} & \textbf{0.613} \\ \midrule
	E.Coli                      & \textbf{0.022} & \textbf{0.022} &     0.011      &     0.009      & \textbf{0.005} & \textbf{0.005} &     0.004      &     0.003      & \textbf{0.684} & \textbf{0.686} & \textbf{0.673} &     0.493      \\ \midrule
	Erdos 971                   &     0.081      & \textbf{0.102} &     0.080      &     0.085      &     0.028      & \textbf{0.037} &     0.028      &     0.031      &     0.757      & \textbf{0.826} &     0.798      &     0.811      \\ \midrule
	Erdos 981                   &     0.083      & \textbf{0.104} &     0.080      &     0.093      &     0.028      & \textbf{0.039} &     0.028      & \textbf{0.037} &     0.758      & \textbf{0.828} &     0.799      &     0.811      \\ \midrule
	Erdos 991                   &     0.079      & \textbf{0.100} &     0.080      &     0.091      &     0.027      & \textbf{0.039} &     0.027      &     0.034      &     0.758      & \textbf{0.828} &     0.801      &     0.810      \\ \midrule
	Everglades                  & \textbf{0.457} & \textbf{0.456} &     0.282      &     0.163      & \textbf{0.389} & \textbf{0.399} &     0.240      &     0.137      &     0.898      & \textbf{0.909} &     0.829      &     0.766      \\ \midrule
	GD 01                       &     0.025      & \textbf{0.089} &     0.058      &     0.070      &     0.014      & \textbf{0.043} &     0.020      & \textbf{0.044} &     0.681      & \textbf{0.805} &     0.752      &     0.786      \\ \midrule
	Haggle Contact              & \textbf{0.590} &     0.563      &     0.533      &     0.424      & \textbf{0.629} &     0.577      &     0.540      &     0.415      &     0.968      & \textbf{0.972} &     0.967      &     0.921      \\ \midrule
	Infectious                  &     0.095      & \textbf{0.208} & \textbf{0.211} & \textbf{0.205} &     0.038      & \textbf{0.120} & \textbf{0.121} &     0.108      &     0.807      & \textbf{0.941} &     0.938      &     0.912      \\ \midrule
	Japan Air                   & \textbf{0.261} & \textbf{0.251} &     0.202      &     0.149      & \textbf{0.223} &     0.201      &     0.170      &     0.087      & \textbf{0.918} & \textbf{0.917} &     0.848      &     0.741      \\ \midrule
	Jazz                        &     0.343      &     0.404      & \textbf{0.433} &     0.391      &     0.283      &     0.379      & \textbf{0.420} &     0.397      &     0.921      & \textbf{0.937} &     0.929      & \textbf{0.936} \\ \midrule
	Les Miserables              &     0.327      & \textbf{0.476} &     0.348      & \textbf{0.494} &     0.226      & \textbf{0.448} &     0.285      & \textbf{0.432} &     0.892      & \textbf{0.951} &     0.846      &     0.905      \\ \midrule
	Macaque Neural              &     0.534      & \textbf{0.591} &     0.490      &     0.512      &     0.505      & \textbf{0.594} &     0.474      &     0.520      &     0.962      & \textbf{0.971} &     0.947      &     0.958      \\ \midrule
	Manufacturing e-mail        &     0.314      & \textbf{0.420} &     0.379      &     0.348      &     0.270      & \textbf{0.413} &     0.385      &     0.326      &     0.905      & \textbf{0.932} &     0.914      &     0.916      \\ \midrule
	Maspalomas                  &     0.163      & \textbf{0.230} &     0.171      &     0.151      &     0.148      & \textbf{0.189} &     0.148      &     0.118      &     0.720      & \textbf{0.771} &     0.720      &     0.691      \\ \midrule
	Narragan                    & \textbf{0.371} &     0.340      &     0.273      &     0.222      & \textbf{0.284} &     0.256      &     0.194      &     0.183      & \textbf{0.824} & \textbf{0.836} &     0.772      &     0.788      \\ \midrule
	Physicians                  &     0.031      & \textbf{0.084} & \textbf{0.078} & \textbf{0.085} &     0.009      & \textbf{0.040} & \textbf{0.039} & \textbf{0.042} &     0.606      & \textbf{0.901} &     0.899      &     0.895      \\ \midrule
	Polbooks                    & \textbf{0.165} & \textbf{0.163} &     0.136      &     0.147      &     0.098      &     0.103      &     0.092      & \textbf{0.114} &     0.812      & \textbf{0.890} &     0.876      & \textbf{0.885} \\ \midrule
	Political Blogs             &     0.107      & \textbf{0.120} &     0.103      & \textbf{0.115} &     0.041      & \textbf{0.053} &     0.046      &     0.045      &     0.849      & \textbf{0.895} & \textbf{0.892} &     0.866      \\ \midrule
	Residence Hall              &     0.087      &     0.154      & \textbf{0.169} &     0.158      &     0.045      &     0.096      & \textbf{0.104} &     0.100      &     0.794      & \textbf{0.860} &     0.842      &     0.844      \\ \midrule
	SFBD Food Web               & \textbf{0.290} &     0.267      &     0.133      &     0.110      & \textbf{0.239} &     0.215      &     0.104      &     0.083      &     0.822      & \textbf{0.854} &     0.792      &     0.755      \\ \midrule
	SFBW Food Web               & \textbf{0.291} &     0.268      &     0.138      &     0.110      & \textbf{0.238} &     0.214      &     0.107      &     0.084      &     0.819      & \textbf{0.854} &     0.795      &     0.759      \\ \midrule
	School                      &     0.225      &     0.324      & \textbf{0.350} &     0.278      &     0.169      &     0.276      & \textbf{0.302} &     0.206      &     0.870      & \textbf{0.895} &     0.889      &     0.879      \\ \midrule
	StMarks                     & \textbf{0.233} &     0.202      &     0.152      &     0.147      & \textbf{0.163} &     0.136      &     0.096      &     0.099      &     0.745      & \textbf{0.765} &     0.668      &     0.720      \\ \midrule
	Terrorist                   &     0.161      & \textbf{0.187} &     0.173      &     0.155      &     0.120      & \textbf{0.146} &     0.125      &     0.113      &     0.803      & \textbf{0.872} &     0.808      &     0.850      \\ \midrule
	Terrorist Train Bombing     &     0.553      & \textbf{0.617} &     0.340      &     0.397      &     0.537      & \textbf{0.627} &     0.292      &     0.362      &     0.833      & \textbf{0.910} &     0.849      & \textbf{0.908} \\ \midrule
	US Air 97                   &     0.348      & \textbf{0.391} &     0.353      &     0.289      &     0.303      & \textbf{0.349} &     0.292      &     0.198      &     0.913      & \textbf{0.955} &     0.917      &     0.935      \\ \midrule
	Zakarays Karate Club        & \textbf{0.212} &     0.158      &     0.152      &     0.066      & \textbf{0.181} &     0.143      &     0.136      &     0.067      &     0.810      & \textbf{0.850} &     0.790      &     0.764      \\ \midrule
	Average significant ranking &     2.550      & \textbf{1.613} &     2.737      &     3.100      &     2.550      & \textbf{1.712} &     2.712      &     3.025      &     2.938      & \textbf{1.262} &     2.925      &     2.875      \\ \bottomrule
\end{tabular}
\end{table*}

\subsection{Detailed results for the experiment of Section \ref{sec:comp}}
\label{sec:app-comp}
Table \ref{tab:comp-alg} contains the per-network TPR, AUPR and AUROC comparison results against algebraic methods as described in Section \ref{sec:comp}. The results of comparison against non-algebraic methods on the networks of Table \ref{tab:data} are shown in Table \ref{tab:nalg-small-tpr} for TPR, Table \ref{tab:nalg-small-aupr} for AUPR, and Table \ref{tab:nalg-small-auroc} for AUROC. The detailed results for the experiment on the larger networks of Table \ref{tab:data-large} are shown in Table \ref{tab:large}.

\begin{table*}[!t]
    \centering
    \caption{Performance results comparison against algebraic embedding methods on networks with less than 1000 nodes. For each performance measure, the best results significant with $p$-value 0.05 are shown in bold. The last row shows the average significant rank of each variant; lower ranks are better.}
    \label{tab:comp-alg}
\begin{tabular}{lccccccccc}
	\toprule
	                            &             \multicolumn{3}{c}{TPR}              &             \multicolumn{3}{c}{AUPR}             &            \multicolumn{3}{c}{AUROC}             \\ 
	                            \cmidrule(lr){2-4} \cmidrule(lr){5-7} \cmidrule(lr){8-10}

	Network                     &      LEM       &      LLE       &      PSL       &      LEM       &      LLE       &      PSL       &      LEM       &      LLE       &      PSL       \\ \midrule
	ACM2009 Contacts            &     0.088      &     0.095      & \textbf{0.229} &     0.076      &     0.080      & \textbf{0.176} &     0.595      &     0.637      & \textbf{0.765} \\ \midrule
	C.Elegans Metabolic         &     0.107      &     0.032      & \textbf{0.164} &     0.058      &     0.012      & \textbf{0.081} & \textbf{0.876} &     0.796      & \textbf{0.874} \\ \midrule
	C.Elegans Neural            &     0.094      &     0.052      & \textbf{0.118} &     0.048      &     0.029      & \textbf{0.064} &     0.863      &     0.825      & \textbf{0.881} \\ \midrule
	CPAN Authors                & \textbf{0.099} &     0.006      & \textbf{0.106} & \textbf{0.040} &     0.002      &     0.037      &     0.869      &     0.661      & \textbf{0.899} \\ \midrule
	Centrality Literature       &     0.144      &     0.030      & \textbf{0.217} &     0.084      &     0.022      & \textbf{0.139} &     0.801      &     0.707      & \textbf{0.877} \\ \midrule
	Chesapeake Lower            &     0.101      &     0.086      & \textbf{0.279} &     0.077      &     0.069      & \textbf{0.231} &     0.681      &     0.676      & \textbf{0.875} \\ \midrule
	Chesapeake Middle           &     0.173      &     0.159      & \textbf{0.355} &     0.143      &     0.123      & \textbf{0.338} &     0.730      &     0.723      & \textbf{0.883} \\ \midrule
	Chesapeake Upper            &     0.118      &     0.111      & \textbf{0.362} &     0.099      &     0.095      & \textbf{0.290} &     0.685      &     0.694      & \textbf{0.867} \\ \midrule
	Codeminer                   &     0.009      &     0.003      & \textbf{0.029} &     0.003      &     0.002      & \textbf{0.006} &     0.626      & \textbf{0.699} &     0.665      \\ \midrule
	Cypress Dry                 &     0.156      &     0.151      & \textbf{0.338} &     0.111      &     0.112      & \textbf{0.279} &     0.764      &     0.779      & \textbf{0.915} \\ \midrule
	Cypress Wet                 &     0.161      &     0.130      & \textbf{0.362} &     0.113      &     0.100      & \textbf{0.301} &     0.763      &     0.774      & \textbf{0.916} \\ \midrule
	DNA Citation                & \textbf{0.027} &     0.012      &     0.008      & \textbf{0.027} &     0.019      &     0.018      &     0.640      & \textbf{0.689} &     0.628      \\ \midrule
	DNA Citation CC             & \textbf{0.023} & \textbf{0.018} & \textbf{0.022} & \textbf{0.029} & \textbf{0.026} & \textbf{0.026} &     0.635      & \textbf{0.693} &     0.624      \\ \midrule
	E.Coli                      & \textbf{0.021} &     0.002      & \textbf{0.022} & \textbf{0.006} &     0.001      & \textbf{0.005} &     0.659      &     0.655      & \textbf{0.684} \\ \midrule
	Erdos 971                   &     0.065      &     0.011      & \textbf{0.100} &     0.018      &     0.005      & \textbf{0.037} &     0.755      &     0.753      & \textbf{0.824} \\ \midrule
	Erdos 981                   &     0.068      &     0.008      & \textbf{0.103} &     0.019      &     0.004      & \textbf{0.040} &     0.759      &     0.753      & \textbf{0.828} \\ \midrule
	Erdos 991                   &     0.065      &     0.009      & \textbf{0.103} &     0.020      &     0.004      & \textbf{0.038} &     0.766      &     0.752      & \textbf{0.830} \\ \midrule
	Everglades                  &     0.143      &     0.146      & \textbf{0.460} &     0.116      &     0.118      & \textbf{0.400} &     0.695      &     0.707      & \textbf{0.911} \\ \midrule
	GD 01                       &     0.014      &     0.028      & \textbf{0.085} &     0.007      &     0.009      & \textbf{0.039} &     0.694      &     0.726      & \textbf{0.804} \\ \midrule
	Haggle Contact              & \textbf{0.590} &     0.053      &     0.564      & \textbf{0.626} &     0.034      &     0.579      &     0.956      &     0.805      & \textbf{0.973} \\ \midrule
	Infectious                  & \textbf{0.203} &     0.154      & \textbf{0.209} & \textbf{0.119} &     0.091      & \textbf{0.121} & \textbf{0.942} &     0.930      & \textbf{0.941} \\ \midrule
	Japan Air                   &     0.204      &     0.096      & \textbf{0.255} &     0.150      &     0.055      & \textbf{0.203} &     0.854      &     0.704      & \textbf{0.916} \\ \midrule
	Jazz                        &     0.375      &     0.194      & \textbf{0.403} &     0.318      &     0.137      & \textbf{0.376} &     0.910      &     0.879      & \textbf{0.937} \\ \midrule
	Les Miserables              &     0.261      &     0.226      & \textbf{0.477} &     0.217      &     0.134      & \textbf{0.449} &     0.880      &     0.842      & \textbf{0.950} \\ \midrule
	Macaque Neural              &     0.390      &     0.385      & \textbf{0.599} &     0.348      &     0.336      & \textbf{0.600} &     0.913      &     0.920      & \textbf{0.972} \\ \midrule
	Manufacturing e-mail        & \textbf{0.414} &     0.104      & \textbf{0.417} & \textbf{0.421} &     0.084      & \textbf{0.411} &     0.919      &     0.760      & \textbf{0.931} \\ \midrule
	Maspalomas                  &     0.085      &     0.054      & \textbf{0.198} &     0.081      &     0.070      & \textbf{0.187} &     0.649      &     0.638      & \textbf{0.776} \\ \midrule
	Narragan                    &     0.196      &     0.197      & \textbf{0.336} &     0.152      &     0.147      & \textbf{0.255} &     0.734      &     0.738      & \textbf{0.839} \\ \midrule
	Physicians                  &     0.071      &     0.053      & \textbf{0.081} &     0.034      &     0.028      & \textbf{0.039} &     0.877      & \textbf{0.903} & \textbf{0.901} \\ \midrule
	Polbooks                    &     0.054      &     0.020      & \textbf{0.166} &     0.040      &     0.030      & \textbf{0.107} &     0.798      &     0.818      & \textbf{0.890} \\ \midrule
	Political Blogs             &     0.073      &     0.015      & \textbf{0.118} &     0.028      &     0.006      & \textbf{0.052} &     0.847      &     0.823      & \textbf{0.895} \\ \midrule
	Residence Hall              &     0.132      &     0.134      & \textbf{0.154} &     0.075      &     0.075      & \textbf{0.093} &     0.785      &     0.813      & \textbf{0.859} \\ \midrule
	SFBD Food Web               &     0.099      &     0.097      & \textbf{0.269} &     0.072      &     0.071      & \textbf{0.214} &     0.692      &     0.697      & \textbf{0.854} \\ \midrule
	SFBW Food Web               &     0.095      &     0.091      & \textbf{0.269} &     0.070      &     0.069      & \textbf{0.217} &     0.692      &     0.700      & \textbf{0.855} \\ \midrule
	School                      & \textbf{0.360} & \textbf{0.371} &     0.320      &     0.290      & \textbf{0.305} &     0.274      &     0.877      &     0.886      & \textbf{0.895} \\ \midrule
	StMarks                     &     0.095      &     0.091      & \textbf{0.207} &     0.062      &     0.058      & \textbf{0.139} &     0.615      &     0.604      & \textbf{0.764} \\ \midrule
	Terrorist                   &     0.141      &     0.038      & \textbf{0.172} &     0.106      &     0.043      & \textbf{0.139} &     0.800      &     0.838      & \textbf{0.864} \\ \midrule
	Terrorist Train Bombing     &     0.455      &     0.183      & \textbf{0.612} &     0.447      &     0.121      & \textbf{0.624} &     0.825      &     0.820      & \textbf{0.910} \\ \midrule
	US Air 97                   &     0.321      &     0.041      & \textbf{0.391} &     0.281      &     0.027      & \textbf{0.347} &     0.906      &     0.837      & \textbf{0.955} \\ \midrule
	Zakarays Karate Club        & \textbf{0.152} &     0.055      & \textbf{0.164} &     0.123      &     0.056      & \textbf{0.146} &     0.814      &     0.741      & \textbf{0.855} \\ \midrule
	Average significant ranking &     2.050      &     2.737      & \textbf{1.212} &     2.050      &     2.750      & \textbf{1.200} &     2.462      &     2.400      & \textbf{1.137} \\ \bottomrule
\end{tabular}
\end{table*}

\begin{table*}[!t]
    \centering
    \caption{TPR results comparison against non-algebraic embedding methods with embedding dimensions 8 and 32 on networks with less than 1000 nodes. For each dimensionlity, the best results significant with $p$-value 0.05 are shown in bold. The last row shows the average significant rank of each variant; lower ranks are better.}
    \label{tab:nalg-small-tpr}
\begin{tabular}{lcccccccccccc}
	\toprule
	                            &                                 \multicolumn{6}{c}{$D=8$}                                  &                                     \multicolumn{6}{c}{$D=32$}                                      \\ \cmidrule(lr){2-7} \cmidrule(lr){8-13}
	Network                     &  DPW  &      LIN       &      MDS       &      MFC       &      N2V       &      PSL       &      DPW       &      LIN       &      MDS       &      MFC       &      N2V       &      PSL       \\ \midrule
	ACM2009 Contacts            & 0.128 &     0.091      &     0.188      & \textbf{0.241} &     0.109      &     0.227      &     0.217      &     0.154      &     0.186      &     0.172      &     0.183      & \textbf{0.226} \\ \midrule
	C.Elegans Metabolic         & 0.072 &     0.081      &     0.086      & \textbf{0.160} &     0.060      & \textbf{0.163} &     0.155      &     0.089      &     0.092      &     0.114      &     0.154      & \textbf{0.213} \\ \midrule
	C.Elegans Neural            & 0.055 &     0.039      &     0.075      &     0.112      &     0.058      & \textbf{0.118} &     0.104      &     0.065      &     0.073      &     0.109      &     0.098      & \textbf{0.145} \\ \midrule
	CPAN Authors                & 0.092 & \textbf{0.108} &     0.094      & \textbf{0.108} &     0.049      & \textbf{0.114} &     0.092      &     0.068      &     0.039      &     0.119      &     0.060      & \textbf{0.131} \\ \midrule
	Centrality Literature       & 0.096 &     0.042      &     0.125      &     0.185      &     0.079      & \textbf{0.225} &     0.196      &     0.120      &     0.125      &     0.137      &     0.135      & \textbf{0.211} \\ \midrule
	Chesapeake Lower            & 0.194 &     0.103      & \textbf{0.302} &     0.265      &     0.181      & \textbf{0.292} &     0.238      &     0.228      & \textbf{0.304} &     0.218      &     0.159      &     0.260      \\ \midrule
	Chesapeake Middle           & 0.290 &     0.127      & \textbf{0.341} & \textbf{0.350} &     0.239      & \textbf{0.359} &     0.279      &     0.265      & \textbf{0.315} &     0.235      &     0.206      & \textbf{0.309} \\ \midrule
	Chesapeake Upper            & 0.253 &     0.165      & \textbf{0.394} &     0.310      &     0.218      &     0.358      &     0.291      &     0.319      & \textbf{0.388} &     0.234      &     0.139      &     0.270      \\ \midrule
	Codeminer                   & 0.009 &     0.003      &     0.004      &     0.018      &     0.009      & \textbf{0.030} & \textbf{0.046} &     0.001      &     0.000      &     0.024      & \textbf{0.040} & \textbf{0.040} \\ \midrule
	Cypress Dry                 & 0.255 &     0.090      &     0.235      &     0.313      &     0.234      & \textbf{0.336} &     0.401      &     0.196      &     0.224      &     0.280      &     0.280      & \textbf{0.450} \\ \midrule
	Cypress Wet                 & 0.271 &     0.089      &     0.225      &     0.309      &     0.228      & \textbf{0.369} &     0.411      &     0.199      &     0.229      &     0.277      &     0.282      & \textbf{0.460} \\ \midrule
	DNA Citation                & 0.012 &     0.010      &     0.023      &     0.012      & \textbf{0.025} &     0.022      & \textbf{0.017} & \textbf{0.027} & \textbf{0.020} & \textbf{0.018} & \textbf{0.022} & \textbf{0.032} \\ \midrule
	DNA Citation CC             & 0.008 &     0.017      & \textbf{0.052} &     0.012      &     0.020      &     0.025      &     0.015      & \textbf{0.050} &     0.037      &     0.020      &     0.012      & \textbf{0.050} \\ \midrule
	E.Coli                      & 0.015 &     0.015      & \textbf{0.024} &     0.015      &     0.010      &     0.019      & \textbf{0.029} &     0.011      &     0.006      &     0.020      &     0.025      &     0.021      \\ \midrule
	Erdos 971                   & 0.027 &     0.004      &     0.005      &     0.089      &     0.028      & \textbf{0.098} &     0.060      &     0.003      &     0.002      &     0.080      &     0.060      & \textbf{0.103} \\ \midrule
	Erdos 981                   & 0.026 &     0.005      &     0.005      &     0.093      &     0.027      & \textbf{0.107} &     0.063      &     0.003      &     0.001      &     0.077      &     0.065      & \textbf{0.106} \\ \midrule
	Erdos 991                   & 0.028 &     0.003      &     0.004      &     0.086      &     0.030      & \textbf{0.098} &     0.061      &     0.004      &     0.002      &     0.077      &     0.060      & \textbf{0.101} \\ \midrule
	Everglades                  & 0.361 &     0.146      &     0.354      &     0.336      &     0.276      & \textbf{0.459} & \textbf{0.456} &     0.290      &     0.357      &     0.328      &     0.335      & \textbf{0.463} \\ \midrule
	GD 01                       & 0.052 &     0.005      &     0.001      & \textbf{0.100} &     0.040      &     0.087      &     0.074      &     0.003      &     0.004      &     0.089      &     0.061      & \textbf{0.115} \\ \midrule
	Haggle Contact              & 0.204 &     0.048      &     0.208      &     0.556      &     0.084      & \textbf{0.570} &     0.439      &     0.177      &     0.290      &     0.422      &     0.311      & \textbf{0.500} \\ \midrule
	Infectious                  & 0.152 &     0.008      &     0.006      & \textbf{0.267} &     0.176      &     0.208      &     0.256      &     0.005      &     0.004      & \textbf{0.310} &     0.227      &     0.296      \\ \midrule
	Japan Air                   & 0.191 &     0.088      &     0.173      &     0.208      &     0.143      & \textbf{0.233} &     0.172      & \textbf{0.200} &     0.164      &     0.111      &     0.106      &     0.179      \\ \midrule
	Jazz                        & 0.223 &     0.033      &     0.072      & \textbf{0.469} &     0.244      &     0.407      &     0.436      &     0.060      &     0.069      &     0.503      &     0.415      & \textbf{0.535} \\ \midrule
	Les Miserables              & 0.364 &     0.046      &     0.079      & \textbf{0.495} &     0.356      & \textbf{0.475} &     0.398      &     0.090      &     0.070      &     0.280      &     0.385      & \textbf{0.446} \\ \midrule
	Macaque Neural              & 0.512 &     0.108      &     0.178      &     0.555      &     0.410      & \textbf{0.593} & \textbf{0.647} &     0.206      &     0.185      &     0.558      &     0.482      &     0.633      \\ \midrule
	Manufacturing e-mail        & 0.213 &     0.086      &     0.276      & \textbf{0.481} &     0.192      &     0.424      &     0.390      &     0.184      &     0.278      &     0.416      &     0.375      & \textbf{0.445} \\ \midrule
	Maspalomas                  & 0.120 &     0.107      & \textbf{0.258} &     0.172      &     0.116      &     0.206      &     0.155      &     0.122      & \textbf{0.320} &     0.130      &     0.120      &     0.170      \\ \midrule
	Narragan                    & 0.289 &     0.205      & \textbf{0.447} &     0.343      &     0.290      &     0.341      &     0.281      &     0.330      & \textbf{0.426} &     0.306      &     0.198      &     0.310      \\ \midrule
	Physicians                  & 0.057 &     0.007      &     0.006      & \textbf{0.078} &     0.060      & \textbf{0.085} &     0.071      &     0.013      &     0.005      &     0.075      &     0.068      & \textbf{0.086} \\ \midrule
	Polbooks                    & 0.086 &     0.021      &     0.016      &     0.137      &     0.081      & \textbf{0.164} & \textbf{0.149} &     0.045      &     0.016      &     0.097      &     0.131      & \textbf{0.155} \\ \midrule
	Political Blogs             & 0.023 &     0.041      &     0.058      & \textbf{0.117} &     0.029      & \textbf{0.123} &     0.074      &     0.013      &     0.004      &     0.095      &     0.052      & \textbf{0.111} \\ \midrule
	Residence Hall              & 0.118 &     0.015      &     0.028      & \textbf{0.167} &     0.132      &     0.153      &     0.166      &     0.020      &     0.018      &     0.202      &     0.168      & \textbf{0.223} \\ \midrule
	SFBD Food Web               & 0.141 &     0.066      &     0.175      &     0.180      &     0.108      & \textbf{0.268} &     0.391      &     0.132      &     0.233      &     0.225      &     0.294      & \textbf{0.440} \\ \midrule
	SFBW Food Web               & 0.129 &     0.066      &     0.176      &     0.188      &     0.105      & \textbf{0.268} &     0.385      &     0.136      &     0.229      &     0.225      &     0.295      & \textbf{0.432} \\ \midrule
	School                      & 0.253 &     0.040      &     0.050      & \textbf{0.360} &     0.303      &     0.322      &     0.337      &     0.056      &     0.046      &     0.349      &     0.361      & \textbf{0.384} \\ \midrule
	StMarks                     & 0.107 &     0.078      & \textbf{0.243} &     0.186      &     0.145      &     0.210      &     0.168      &     0.184      & \textbf{0.243} &     0.129      &     0.091      &     0.171      \\ \midrule
	Terrorist                   & 0.127 &     0.029      &     0.042      & \textbf{0.186} &     0.122      & \textbf{0.180} & \textbf{0.135} &     0.089      &     0.043      &     0.068      &     0.128      &     0.107      \\ \midrule
	Terrorist Train Bombing     & 0.356 &     0.066      &     0.030      &     0.575      &     0.260      & \textbf{0.616} & \textbf{0.380} &     0.180      &     0.033      &     0.223      &     0.344      &     0.370      \\ \midrule
	US Air 97                   & 0.115 &     0.042      &     0.078      &     0.386      &     0.098      & \textbf{0.395} &     0.313      &     0.083      &     0.113      &     0.327      &     0.264      & \textbf{0.388} \\ \midrule
	Zakarays Karate Club        & 0.124 &     0.075      &     0.139      & \textbf{0.160} &     0.135      & \textbf{0.156} & \textbf{0.169} &     0.114      &     0.106      &     0.125      &     0.122      & \textbf{0.188} \\ \midrule
	Average significant ranking & 3.913 &     5.500      &     3.763      &     2.062      &     4.225      & \textbf{1.538} &     2.688      &     4.850      &     4.525      &     3.425      &     3.950      & \textbf{1.562} \\ \bottomrule
\end{tabular}
\end{table*}

\begin{table*}[!t]
    \centering
    \caption{AUPR results comparison against non-algebraic embedding methods with embedding dimensions 8 and 32 on networks with less than 1000 nodes. For each dimensionlity, the best results significant with $p$-value 0.05 are shown in bold. The last row shows the average significant rank of each variant; lower ranks are better.}
    \label{tab:nalg-small-aupr}
\begin{tabular}{lcccccccccccc}
	\toprule
	                            &                                 \multicolumn{6}{c}{$D=8$}                                  &                                     \multicolumn{6}{c}{$D=32$}                                      \\ \cmidrule(lr){2-7} \cmidrule(lr){8-13}
	Network                     &  DPW  &      LIN       &      MDS       &      MFC       &      N2V       &      PSL       &      DPW       &      LIN       &      MDS       &      MFC       &      N2V       &      PSL       \\ \midrule
	ACM2009 Contacts            & 0.098 &     0.071      &     0.131      & \textbf{0.195} &     0.084      &     0.176      & \textbf{0.172} &     0.114      &     0.138      &     0.125      &     0.136      & \textbf{0.171} \\ \midrule
	C.Elegans Metabolic         & 0.028 &     0.019      &     0.026      & \textbf{0.085} &     0.024      & \textbf{0.082} &     0.075      &     0.027      &     0.031      &     0.060      &     0.078      & \textbf{0.128} \\ \midrule
	C.Elegans Neural            & 0.028 &     0.011      &     0.021      &     0.054      &     0.032      & \textbf{0.065} &     0.053      &     0.019      &     0.023      &     0.054      &     0.051      & \textbf{0.081} \\ \midrule
	CPAN Authors                & 0.032 & \textbf{0.038} &     0.030      & \textbf{0.038} &     0.013      & \textbf{0.041} &     0.032      &     0.025      &     0.014      & \textbf{0.047} &     0.015      &     0.039      \\ \midrule
	Centrality Literature       & 0.057 &     0.024      &     0.089      &     0.120      &     0.046      & \textbf{0.141} &     0.123      &     0.067      &     0.095      &     0.085      &     0.083      & \textbf{0.143} \\ \midrule
	Chesapeake Lower            & 0.159 &     0.085      &     0.208      & \textbf{0.235} &     0.149      & \textbf{0.241} &     0.178      &     0.163      & \textbf{0.247} &     0.174      &     0.145      & \textbf{0.248} \\ \midrule
	Chesapeake Middle           & 0.245 &     0.102      &     0.252      & \textbf{0.317} &     0.214      & \textbf{0.335} &     0.253      &     0.210      &     0.248      &     0.211      &     0.197      & \textbf{0.301} \\ \midrule
	Chesapeake Upper            & 0.197 &     0.125      &     0.251      &     0.249      &     0.177      & \textbf{0.297} &     0.244      &     0.230      & \textbf{0.290} &     0.184      &     0.126      &     0.232      \\ \midrule
	Codeminer                   & 0.004 &     0.001      &     0.001      &     0.002      &     0.004      & \textbf{0.006} & \textbf{0.012} &     0.001      &     0.000      &     0.007      & \textbf{0.012} & \textbf{0.009} \\ \midrule
	Cypress Dry                 & 0.197 &     0.064      &     0.159      &     0.254      &     0.201      & \textbf{0.283} &     0.350      &     0.134      &     0.205      &     0.241      &     0.234      & \textbf{0.421} \\ \midrule
	Cypress Wet                 & 0.208 &     0.062      &     0.152      &     0.250      &     0.189      & \textbf{0.305} &     0.360      &     0.134      &     0.216      &     0.227      &     0.238      & \textbf{0.446} \\ \midrule
	DNA Citation                & 0.020 &     0.016      &     0.019      &     0.016      & \textbf{0.028} &     0.021      & \textbf{0.021} & \textbf{0.024} &     0.015      &     0.021      & \textbf{0.024} & \textbf{0.028} \\ \midrule
	DNA Citation CC             & 0.019 &     0.025      & \textbf{0.045} &     0.019      &     0.025      &     0.025      &     0.026      &     0.041      &     0.030      &     0.022      &     0.023      & \textbf{0.049} \\ \midrule
	E.Coli                      & 0.004 &     0.004      & \textbf{0.008} &     0.004      &     0.003      &     0.005      &     0.007      &     0.004      &     0.002      & \textbf{0.009} &     0.007      &     0.007      \\ \midrule
	Erdos 971                   & 0.011 &     0.002      &     0.002      &     0.033      &     0.011      & \textbf{0.037} &     0.020      &     0.002      &     0.001      &     0.027      &     0.022      & \textbf{0.040} \\ \midrule
	Erdos 981                   & 0.010 &     0.002      &     0.002      &     0.032      &     0.011      & \textbf{0.041} &     0.021      &     0.002      &     0.001      &     0.027      &     0.022      & \textbf{0.042} \\ \midrule
	Erdos 991                   & 0.011 &     0.002      &     0.002      &     0.030      &     0.012      & \textbf{0.037} &     0.021      &     0.002      &     0.001      &     0.027      &     0.020      & \textbf{0.039} \\ \midrule
	Everglades                  & 0.303 &     0.109      &     0.288      &     0.284      &     0.223      & \textbf{0.401} &     0.412      &     0.223      &     0.311      &     0.274      &     0.286      & \textbf{0.440} \\ \midrule
	GD 01                       & 0.022 &     0.003      &     0.002      & \textbf{0.039} &     0.019      & \textbf{0.039} &     0.028      &     0.003      &     0.002      &     0.035      &     0.026      & \textbf{0.049} \\ \midrule
	Haggle Contact              & 0.134 &     0.022      &     0.127      & \textbf{0.573} &     0.049      & \textbf{0.582} &     0.430      &     0.105      &     0.249      &     0.370      &     0.276      & \textbf{0.474} \\ \midrule
	Infectious                  & 0.083 &     0.004      &     0.004      & \textbf{0.168} &     0.098      &     0.119      &     0.152      &     0.004      &     0.003      & \textbf{0.213} &     0.135      & \textbf{0.208} \\ \midrule
	Japan Air                   & 0.136 &     0.059      &     0.135      &     0.155      &     0.114      & \textbf{0.193} &     0.132      & \textbf{0.154} &     0.135      &     0.095      &     0.096      &     0.145      \\ \midrule
	Jazz                        & 0.166 &     0.022      &     0.031      & \textbf{0.451} &     0.183      &     0.379      &     0.413      &     0.033      &     0.033      &     0.513      &     0.398      & \textbf{0.555} \\ \midrule
	Les Miserables              & 0.312 &     0.026      &     0.034      & \textbf{0.481} &     0.303      &     0.440      &     0.354      &     0.051      &     0.032      &     0.228      &     0.353      & \textbf{0.406} \\ \midrule
	Macaque Neural              & 0.497 &     0.086      &     0.124      &     0.573      &     0.386      & \textbf{0.596} & \textbf{0.675} &     0.156      &     0.131      &     0.548      &     0.469      & \textbf{0.664} \\ \midrule
	Manufacturing e-mail        & 0.161 &     0.057      &     0.224      & \textbf{0.485} &     0.136      &     0.415      &     0.387      &     0.132      &     0.254      &     0.399      &     0.357      & \textbf{0.449} \\ \midrule
	Maspalomas                  & 0.107 &     0.083      & \textbf{0.178} & \textbf{0.156} &     0.123      & \textbf{0.181} &     0.125      &     0.100      & \textbf{0.208} &     0.112      &     0.123      &     0.150      \\ \midrule
	Narragan                    & 0.218 &     0.144      & \textbf{0.302} & \textbf{0.281} &     0.243      &     0.257      &     0.221      &     0.218      & \textbf{0.359} &     0.253      &     0.173      &     0.283      \\ \midrule
	Physicians                  & 0.031 &     0.004      &     0.004      & \textbf{0.038} &     0.030      & \textbf{0.040} &     0.035      &     0.006      &     0.004      &     0.038      &     0.036      & \textbf{0.044} \\ \midrule
	Polbooks                    & 0.063 &     0.015      &     0.013      &     0.083      &     0.057      & \textbf{0.105} & \textbf{0.089} &     0.024      &     0.013      &     0.063      & \textbf{0.089} & \textbf{0.096} \\ \midrule
	Political Blogs             & 0.009 &     0.007      &     0.010      & \textbf{0.052} &     0.010      & \textbf{0.054} &     0.023      &     0.003      &     0.002      &     0.039      &     0.016      & \textbf{0.047} \\ \midrule
	Residence Hall              & 0.064 &     0.010      &     0.011      & \textbf{0.100} &     0.072      &     0.093      &     0.100      &     0.012      &     0.010      &     0.138      &     0.103      & \textbf{0.150} \\ \midrule
	SFBD Food Web               & 0.105 &     0.049      &     0.123      &     0.133      &     0.083      & \textbf{0.215} &     0.360      &     0.086      &     0.193      &     0.169      &     0.254      & \textbf{0.422} \\ \midrule
	SFBW Food Web               & 0.096 &     0.049      &     0.122      &     0.136      &     0.083      & \textbf{0.214} &     0.362      &     0.088      &     0.187      &     0.170      &     0.255      & \textbf{0.414} \\ \midrule
	School                      & 0.184 &     0.031      &     0.032      & \textbf{0.317} &     0.234      &     0.273      &     0.271      &     0.037      &     0.031      &     0.285      &     0.296      & \textbf{0.334} \\ \midrule
	StMarks                     & 0.078 &     0.056      & \textbf{0.152} &     0.146      &     0.107      &     0.138      &     0.115      &     0.108      & \textbf{0.158} &     0.104      &     0.075      &     0.142      \\ \midrule
	Terrorist                   & 0.091 &     0.023      &     0.028      & \textbf{0.146} &     0.082      & \textbf{0.148} & \textbf{0.102} &     0.048      &     0.028      &     0.053      & \textbf{0.101} &     0.084      \\ \midrule
	Terrorist Train Bombing     & 0.310 &     0.044      &     0.032      &     0.562      &     0.209      & \textbf{0.633} & \textbf{0.369} &     0.130      &     0.030      &     0.194      &     0.309      & \textbf{0.349} \\ \midrule
	US Air 97                   & 0.062 &     0.013      &     0.034      &     0.327      &     0.049      & \textbf{0.352} &     0.239      &     0.036      &     0.062      &     0.258      &     0.193      & \textbf{0.317} \\ \midrule
	Zakarays Karate Club        & 0.107 &     0.058      &     0.081      &     0.137      &     0.120      & \textbf{0.149} & \textbf{0.133} &     0.094      &     0.079      &     0.111      &     0.112      & \textbf{0.152} \\ \midrule
	Average significant ranking & 3.712 &     5.537      &     4.162      &     2.038      &     3.987      & \textbf{1.562} &     2.763      &     4.938      &     4.625      &     3.450      &     3.788      & \textbf{1.438} \\ \bottomrule 
\end{tabular}
\end{table*}

\begin{table*}[!t]
    \centering
    \caption{AUROC results comparison against non-algebraic embedding methods with embedding dimensions 8 and 32 on networks with less than 1000 nodes. For each dimensionlity, the best results significant with $p$-value 0.05 are shown in bold. The last row shows the average significant rank of each variant; lower ranks are better.}
    \label{tab:nalg-small-auroc}
\begin{tabular}{lcccccccccccc}
	\toprule
	                            &                                 \multicolumn{6}{c}{$D=8$}                                  &                                     \multicolumn{6}{c}{$D=32$}                                      \\ \cmidrule(lr){2-7} \cmidrule(lr){8-13}
	Network                     &      DPW       &  LIN  &  MDS  &      MFC       &      N2V       &      PSL       &      DPW       &  LIN  &  MDS  &      MFC       &      N2V       &      PSL       \\ \midrule
	ACM2009 Contacts            &     0.673      & 0.583 & 0.595 & \textbf{0.762} &     0.618      & \textbf{0.765} &     0.734      & 0.660 & 0.664 &     0.696      &     0.704      & \textbf{0.746} \\ \midrule
	C.Elegans Metabolic         &     0.851      & 0.690 & 0.681 &     0.830      &     0.848      & \textbf{0.872} &     0.854      & 0.723 & 0.718 &     0.842      & \textbf{0.876} &     0.863      \\ \midrule
	C.Elegans Neural            &     0.829      & 0.584 & 0.616 &     0.849      &     0.834      & \textbf{0.879} &     0.843      & 0.633 & 0.653 &     0.859      &     0.841      & \textbf{0.884} \\ \midrule
	CPAN Authors                &     0.828      & 0.809 & 0.816 & \textbf{0.901} &     0.721      & \textbf{0.905} &     0.853      & 0.709 & 0.610 & \textbf{0.918} &     0.761      &     0.809      \\ \midrule
	Centrality Literature       &     0.817      & 0.631 & 0.750 &     0.852      &     0.790      & \textbf{0.879} &     0.851      & 0.740 & 0.754 &     0.836      &     0.844      & \textbf{0.862} \\ \midrule
	Chesapeake Lower            &     0.815      & 0.646 & 0.768 &     0.860      &     0.807      & \textbf{0.873} & \textbf{0.804} & 0.769 & 0.779 &     0.754      & \textbf{0.810} & \textbf{0.806} \\ \midrule
	Chesapeake Middle           &     0.836      & 0.649 & 0.755 &     0.869      &     0.828      & \textbf{0.886} & \textbf{0.835} & 0.770 & 0.786 &     0.786      &     0.819      & \textbf{0.840} \\ \midrule
	Chesapeake Upper            &     0.790      & 0.655 & 0.765 &     0.839      &     0.781      & \textbf{0.867} & \textbf{0.813} & 0.735 & 0.763 &     0.742      &     0.762      &     0.780      \\ \midrule
	Codeminer                   & \textbf{0.708} & 0.588 & 0.525 &     0.568      &     0.659      &     0.672      &     0.649      & 0.505 & 0.500 &     0.644      & \textbf{0.683} &     0.633      \\ \midrule
	Cypress Dry                 &     0.861      & 0.631 & 0.672 &     0.901      &     0.843      & \textbf{0.916} &     0.915      & 0.745 & 0.741 &     0.823      &     0.874      & \textbf{0.921} \\ \midrule
	Cypress Wet                 &     0.860      & 0.623 & 0.664 &     0.899      &     0.835      & \textbf{0.917} &     0.917      & 0.739 & 0.774 &     0.814      &     0.878      & \textbf{0.926} \\ \midrule
	DNA Citation                & \textbf{0.667} & 0.545 & 0.509 &     0.580      & \textbf{0.670} &     0.632      & \textbf{0.653} & 0.587 & 0.503 &     0.586      & \textbf{0.662} &     0.585      \\ \midrule
	DNA Citation CC             &     0.616      & 0.534 & 0.543 &     0.604      & \textbf{0.638} &     0.628      &     0.616      & 0.564 & 0.581 &     0.589      & \textbf{0.651} &     0.631      \\ \midrule
	E.Coli                      & \textbf{0.700} & 0.637 & 0.624 &     0.589      &     0.694      &     0.682      &     0.678      & 0.613 & 0.507 & \textbf{0.786} &     0.663      &     0.712      \\ \midrule
	Erdos 971                   &     0.808      & 0.557 & 0.513 &     0.758      &     0.808      & \textbf{0.823} &     0.777      & 0.520 & 0.498 &     0.693      &     0.775      & \textbf{0.803} \\ \midrule
	Erdos 981                   &     0.808      & 0.560 & 0.518 &     0.755      &     0.812      & \textbf{0.828} &     0.778      & 0.519 & 0.500 &     0.699      &     0.780      & \textbf{0.809} \\ \midrule
	Erdos 991                   &     0.811      & 0.555 & 0.518 &     0.753      &     0.813      & \textbf{0.831} &     0.777      & 0.519 & 0.497 &     0.696      &     0.783      & \textbf{0.810} \\ \midrule
	Everglades                  &     0.867      & 0.632 & 0.812 &     0.844      &     0.785      & \textbf{0.911} & \textbf{0.898} & 0.755 & 0.832 &     0.794      &     0.843      &     0.890      \\ \midrule
	GD 01                       & \textbf{0.799} & 0.542 & 0.509 &     0.760      & \textbf{0.799} & \textbf{0.806} & \textbf{0.779} & 0.535 & 0.498 &     0.696      & \textbf{0.778} & \textbf{0.784} \\ \midrule
	Haggle Contact              &     0.923      & 0.673 & 0.775 & \textbf{0.975} &     0.843      &     0.973      &     0.959      & 0.837 & 0.905 & \textbf{0.965} &     0.928      & \textbf{0.964} \\ \midrule
	Infectious                  &     0.932      & 0.540 & 0.510 &     0.927      &     0.935      & \textbf{0.941} &     0.915      & 0.517 & 0.501 &     0.916      &     0.927      & \textbf{0.945} \\ \midrule
	Japan Air                   &     0.882      & 0.721 & 0.822 & \textbf{0.910} &     0.866      & \textbf{0.915} &     0.877      & 0.870 & 0.778 & \textbf{0.891} &     0.858      & \textbf{0.900} \\ \midrule
	Jazz                        &     0.896      & 0.569 & 0.534 & \textbf{0.939} &     0.901      &     0.937      &     0.939      & 0.640 & 0.586 &     0.940      &     0.940      & \textbf{0.960} \\ \midrule
	Les Miserables              &     0.907      & 0.637 & 0.522 &     0.936      &     0.904      & \textbf{0.951} & \textbf{0.907} & 0.733 & 0.573 &     0.885      & \textbf{0.914} & \textbf{0.909} \\ \midrule
	Macaque Neural              &     0.954      & 0.622 & 0.610 &     0.966      &     0.933      & \textbf{0.972} & \textbf{0.975} & 0.748 & 0.645 &     0.954      &     0.956      &     0.973      \\ \midrule
	Manufacturing e-mail        &     0.845      & 0.622 & 0.696 & \textbf{0.936} &     0.796      &     0.932      &     0.913      & 0.739 & 0.814 &     0.913      &     0.909      & \textbf{0.930} \\ \midrule
	Maspalomas                  &     0.703      & 0.595 & 0.719 &     0.729      &     0.725      & \textbf{0.775} &     0.697      & 0.619 & 0.701 &     0.669      & \textbf{0.731} &     0.677      \\ \midrule
	Narragan                    &     0.805      & 0.684 & 0.749 &     0.794      &     0.801      & \textbf{0.836} & \textbf{0.793} & 0.774 & 0.760 &     0.767      &     0.769      &     0.776      \\ \midrule
	Physicians                  & \textbf{0.902} & 0.542 & 0.529 &     0.887      &     0.897      & \textbf{0.900} &     0.881      & 0.569 & 0.507 &     0.867      & \textbf{0.894} &     0.882      \\ \midrule
	Polbooks                    &     0.871      & 0.591 & 0.538 &     0.862      &     0.873      & \textbf{0.891} & \textbf{0.883} & 0.676 & 0.547 &     0.837      & \textbf{0.890} &     0.873      \\ \midrule
	Political Blogs             &     0.836      & 0.631 & 0.632 &     0.887      &     0.830      & \textbf{0.896} &     0.784      & 0.535 & 0.501 & \textbf{0.844} &     0.768      & \textbf{0.852} \\ \midrule
	Residence Hall              &     0.820      & 0.534 & 0.510 &     0.850      &     0.832      & \textbf{0.859} &     0.834      & 0.556 & 0.500 &     0.841      &     0.840      & \textbf{0.868} \\ \midrule
	SFBD Food Web               &     0.766      & 0.573 & 0.650 &     0.805      &     0.751      & \textbf{0.854} &     0.895      & 0.636 & 0.785 &     0.831      &     0.863      & \textbf{0.917} \\ \midrule
	SFBW Food Web               &     0.761      & 0.574 & 0.652 &     0.807      &     0.753      & \textbf{0.853} &     0.895      & 0.637 & 0.780 &     0.831      &     0.865      & \textbf{0.912} \\ \midrule
	School                      &     0.852      & 0.538 & 0.522 & \textbf{0.893} &     0.868      & \textbf{0.894} &     0.878      & 0.580 & 0.515 &     0.871      &     0.884      & \textbf{0.894} \\ \midrule
	StMarks                     &     0.690      & 0.595 & 0.662 &     0.738      &     0.708      & \textbf{0.768} & \textbf{0.731} & 0.690 & 0.663 &     0.642      &     0.681      & \textbf{0.734} \\ \midrule
	Terrorist                   &     0.845      & 0.617 & 0.509 &     0.856      &     0.853      & \textbf{0.871} & \textbf{0.845} & 0.704 & 0.549 &     0.757      & \textbf{0.856} &     0.812      \\ \midrule
	Terrorist Train Bombing     &     0.893      & 0.660 & 0.637 &     0.885      &     0.888      & \textbf{0.908} & \textbf{0.897} & 0.741 & 0.625 &     0.816      & \textbf{0.896} &     0.861      \\ \midrule
	US Air 97                   &     0.905      & 0.643 & 0.668 &     0.936      &     0.885      & \textbf{0.955} &     0.938      & 0.761 & 0.793 &     0.925      &     0.929      & \textbf{0.944} \\ \midrule
	Zakarays Karate Club        &     0.820      & 0.665 & 0.727 &     0.801      &     0.829      & \textbf{0.850} & \textbf{0.830} & 0.791 & 0.735 &     0.791      &     0.807      &     0.802      \\ \midrule
	Average significant ranking &     2.925      & 5.475 & 5.338 &     2.700      &     3.237      & \textbf{1.325} &     2.250      & 5.188 & 5.375 &     3.688      &     2.688      & \textbf{1.812} \\ \bottomrule 
\end{tabular}
\end{table*}

\begin{table*}[!t]
    \centering
    \caption{Performance results comparison on networks with more than 1000 nodes. The best results significant with $p$-value 0.05 are shown in bold. The last row shows the average significant rank of each variant; lower ranks are better.}
    \label{tab:large}
\begin{tabular}{lcccccccccccc}
	\toprule
	                            &                 \multicolumn{4}{c}{TPR}                  &            \multicolumn{4}{c}{AUPR}             &                     \multicolumn{4}{c}{AUROC}                     \\
	                            \cmidrule(lr){2-5} \cmidrule(lr){6-9} \cmidrule(lr){10-13}
	Network                     &      DPW       &      MFC       &  N2V  &      PSL       &  DPW  &      MFC       &  N2V  &      PSL       &      DPW       &      MFC       &      N2V       &      PSL       \\ \midrule
	Adolescent Health           &     0.409      &     0.328      & 0.402 & \textbf{0.459} & 0.388 &     0.309      & 0.376 & \textbf{0.449} &     0.734      &     0.593      &     0.726      & \textbf{0.746} \\ \midrule
	Advogato                    &     0.349      &     0.636      & 0.446 & \textbf{0.716} & 0.315 &     0.624      & 0.393 & \textbf{0.781} &     0.654      &     0.859      &     0.729      & \textbf{0.901} \\ \midrule
	BitcoinAlpha                &     0.162      &     0.554      & 0.243 & \textbf{0.681} & 0.191 &     0.509      & 0.229 & \textbf{0.737} &     0.618      &     0.844      &     0.668      & \textbf{0.895} \\ \midrule
	Ciao                        &     0.451      &     0.722      & 0.511 & \textbf{0.773} & 0.387 &     0.668      & 0.448 & \textbf{0.840} &     0.758      &     0.897      &     0.788      & \textbf{0.934} \\ \midrule
	Criminal                    &     0.114      &     0.348      & 0.092 & \textbf{0.485} & 0.114 &     0.274      & 0.076 & \textbf{0.422} &     0.667      & \textbf{0.715} &     0.461      &     0.662      \\ \midrule
	DNC Email                   &     0.500      &     0.744      & 0.465 & \textbf{0.795} & 0.430 &     0.755      & 0.409 & \textbf{0.848} &     0.792      &     0.927      &     0.743      & \textbf{0.937} \\ \midrule
	Diseasome                   &     0.163      &     0.397      & 0.176 & \textbf{0.461} & 0.224 &     0.377      & 0.202 & \textbf{0.465} &     0.567      &     0.692      &     0.490      & \textbf{0.723} \\ \midrule
	FAA                         &     0.323      &     0.371      & 0.336 & \textbf{0.453} & 0.310 &     0.358      & 0.328 & \textbf{0.452} &     0.649      &     0.610      & \textbf{0.673} &     0.640      \\ \midrule
	Facebook                    & \textbf{0.858} &     0.710      & 0.848 &     0.837      & 0.836 &     0.681      & 0.828 & \textbf{0.846} & \textbf{0.909} &     0.737      &     0.898      &     0.888      \\ \midrule
	GR                          &     0.218      &     0.296      & 0.165 & \textbf{0.462} & 0.167 &     0.209      & 0.159 & \textbf{0.421} &     0.705      &     0.714      &     0.699      & \textbf{0.751} \\ \midrule
	Hero                        &     0.518      &     0.764      & 0.539 & \textbf{0.802} & 0.451 &     0.812      & 0.482 & \textbf{0.878} &     0.776      &     0.935      &     0.795      & \textbf{0.948} \\ \midrule
	Human Protein               &     0.190      &     0.441      & 0.241 & \textbf{0.555} & 0.195 &     0.364      & 0.207 & \textbf{0.557} &     0.557      &     0.739      &     0.530      & \textbf{0.768} \\ \midrule
	Indochina 2004              &     0.130      &     0.281      & 0.130 & \textbf{0.404} & 0.138 &     0.219      & 0.131 & \textbf{0.321} &     0.769      &     0.763      &     0.743      & \textbf{0.787} \\ \midrule
	ODLIS                       &     0.313      &     0.589      & 0.496 & \textbf{0.696} & 0.304 &     0.578      & 0.464 & \textbf{0.772} &     0.569      &     0.796      &     0.713      & \textbf{0.864} \\ \midrule
	PGP                         &     0.174      &     0.230      & 0.196 & \textbf{0.386} & 0.096 &     0.141      & 0.126 & \textbf{0.311} &     0.632      & \textbf{0.733} &     0.615      & \textbf{0.736} \\ \midrule
	Roget                       &     0.538      &     0.485      & 0.545 & \textbf{0.597} & 0.497 &     0.494      & 0.514 & \textbf{0.592} &     0.651      &     0.584      &     0.665      & \textbf{0.704} \\ \midrule
	Web EPA                     &     0.097      & \textbf{0.499} & 0.149 &     0.478      & 0.121 & \textbf{0.453} & 0.130 & \textbf{0.471} &     0.644      & \textbf{0.892} &     0.634      &     0.881      \\ \midrule
	Web Edu                     &     0.242      &     0.387      & 0.191 & \textbf{0.502} & 0.230 &     0.289      & 0.186 & \textbf{0.413} & \textbf{0.761} &     0.696      &     0.686      &     0.666      \\ \midrule
	WikiTalk                    &     0.321      & \textbf{0.852} & 0.656 & \textbf{0.854} & 0.289 & \textbf{0.892} & 0.664 &     0.872      &     0.782      & \textbf{0.973} &     0.901      &     0.967      \\ \midrule
	Yeast                       &     0.224      &     0.436      & 0.274 & \textbf{0.573} & 0.227 &     0.371      & 0.257 & \textbf{0.578} &     0.588      &     0.740      &     0.628      & \textbf{0.806} \\ \midrule
	Youtube                     &     0.272      &     0.394      & 0.234 & \textbf{0.464} & 0.184 &     0.315      & 0.163 & \textbf{0.426} &     0.683      &     0.833      &     0.668      & \textbf{0.849} \\ \midrule
	Average significant ranking &     3.405      &     2.214      & 3.214 & \textbf{1.167} & 3.429 &     2.167      & 3.333 & \textbf{1.071} &     3.024      &     2.286      &     3.190      & \textbf{1.500} \\ \bottomrule
\end{tabular}
\end{table*}

\ifCLASSOPTIONcompsoc
  \section*{Acknowledgments}
        This research work is supported by the Research Center, CCIS, King Saud University, Riyadh, Saudi Arabia.
\else
  \section*{Acknowledgment}
\fi

\ifCLASSOPTIONcaptionsoff
  \newpage
\fi



\bibliographystyle{IEEEtran}
%

%




\end{document}